\documentclass[11pt]{article}

\usepackage[final]{acl}

\usepackage{times}
\usepackage{latexsym}

\usepackage[T1]{fontenc}

\usepackage[utf8]{inputenc}

\usepackage{microtype}

\usepackage{inconsolata}

\usepackage{graphicx}
\usepackage[most]{tcolorbox}

\usepackage{amsmath}
\usepackage[table]{xcolor}
\usepackage{multirow}
\usepackage{tikz}
\usepackage{booktabs}
\usepackage{diagbox}        
\usepackage{booktabs}       
\usepackage{amsfonts}       
\usepackage{nicefrac}       
\usepackage{microtype}      

\usepackage{listings}
\usepackage{graphicx}
\usepackage{pifont}

\usetikzlibrary{arrows.meta, positioning, calc, fit, backgrounds}

\definecolor{tabred}{RGB}{214,39,40}
\definecolor{tabblue}{RGB}{31,119,180}

\usepackage[capitalise]{cleveref}
\usepackage[all]{foreign}

\usepackage{xurl}

\DeclareMathOperator{\V}{\mathcal{V}}
\DeclareMathOperator{\C}{\mathcal{C}}
\DeclareMathOperator{\R}{\mathcal{R}}
\DeclareMathOperator{\F}{\mathcal{F}}

\definecolor{jsonkey}{RGB}{127,0,85}      
\definecolor{jsonstring}{RGB}{0,102,153}  
\definecolor{jsonpunct}{RGB}{80,80,80}    

\lstdefinelanguage{json}{
  basicstyle=\ttfamily\footnotesize,
  showstringspaces=false,
  breaklines=true,
  frame=single,
  rulecolor=\color{black!20},
  columns=fullflexible,
  keepspaces=true,
  upquote=true,
  stringstyle=\color{jsonstring},
  morestring=[b]",
  keywords={
    "id":,
    "NL_sentence":,
    "FOL_sentence":,
    "FOL_sentence_old":,
    "FOL_sentence_new":,
    "label":,
    "label_old":,
    "corrected":,
    "ambiguity":,
    "ambiguity_explanation":,
    "correction_explanation":
  },
  keywordstyle=\color{jsonkey}\bfseries,
  literate=
   {∀}{{$\forall$}}1
   {∃}{{$\exists$}}1
   {∧}{{$\land$}}1
   {∨}{{$\lor$}}1
   {¬}{{$\lnot$}}1
   {→}{{$\rightarrow$}}1
   {↔}{{$\leftrightarrow$}}1
   {⊕}{{$\oplus$}}1
   {≠}{{$\n eq$}}1
   {:}{{{\color{jsonpunct}:}}}1
   {,}{{{\color{jsonpunct},}}}1
   {\{}{{{\color{jsonpunct}\{}}}1
   {\}}{{{\color{jsonpunct}\}}}}1
   {[}{{{\color{jsonpunct}[}}}1
   {]}{{{\color{jsonpunct}]}}}1
}

\lstnewenvironment{jsonblock}{\lstset{language=json}}{}

    \title{Fixing FOLIO and MALLS: \\ Verified Annotations and an LLM-assisted Framework \\ to Focus Human Relabeling}

\author{
  Andrea Brunello$^1$ \quad
  Cristian Curaba$^1$ \quad
  Luca Geatti$^1$ \AND
  Michele Mignani$^{1*}$ \quad
  Angelo Montanari$^1$ \quad
  Nicola Saccomanno$^1$ \\[0.5em]
  $^1$University of Udine, Italy \\
  \texttt{name.surname@uniud.it} \\[0.3em]
  {\small $^*$Corresponding author.}
}

\begin{document}
\maketitle

\begin{abstract}
Accurate translation from Natural Language to First-Order Logic
(NL-to-FOL) underpins neurosymbolic AI systems and Natural Language
Inference (NLI), making the quality of NL-to-FOL benchmarks essential---yet these datasets have never been rigorously audited. Our first contribution is to present
a systematic human inspection of the validation split of \textsf{FOLIO}
and a subset of \textsf{MALLS} test instances, finding that
approximately 42.5\% and 42\% of entries, respectively, contain incorrect
FOL formalizations (i.e., ground truth labels), with additional rates of ambiguous NL sentences
(17.8\% and 51\%) and incorrect NLI labels in \textsf{FOLIO} (8.4\%).
Our second contribution is to develop and release corrected ground truths for such datasets, showing that
annotation errors distort model evaluation on a reference benchmark task: testing three
state-of-the-art LLMs (Gemma~4 31B-it, Qwen3-30B-A3B, and GPT-4o-mini)
with the corrected ground truths yields accuracy gains from +11 to +23
percentage points. 
Motivated by these findings, we propose an LLM-based framework to support humans in manual reviewing NL-to-FOL datasets. By directing reviewers toward the most error-prone instances, we empirically show that it is possible to achieve 90\% dataset accuracy after reviewing fewer than 20\% of instances, compared to over 76\% required by
unguided review. We release all human-verified annotations and the
code for our framework.

\end{abstract}

\section{Introduction}
\label{sec:introduction}

Automatically translating Natural Language (NL) into a machine-readable formalism---often called \textit{autoformalization}---is a fundamental building block of neurosymbolic AI, with applications ranging from Natural Language Inference (NLI) \citep{DBLP:conf/nips/YeCDD23,DBLP:conf/emnlp/OlaussonGLZSTL23,DBLP:conf/emnlp/PanAWW23} to runtime verification and AI safety \citep{Toward_guaranteed_safe_AI}. Among the 
target formalisms, First-Order Logic (FOL) stands out for its expressiveness and computational tractability. Yet, NL-to-FOL translation\footnote{See Appendix~\ref{appendix:task-def}
for a more formal definition.} remains a longstanding challenge for both humans and automated systems \citep{barker2009difficulty,singh2020exploring}. We refer the reader to \citet{fol-survey} for a comprehensive survey of approaches, datasets, and open challenges in NL-to-FOL formalization, and we will adopt the terminology introduced therein throughout this paper.

To support progress in this area, datasets have been developed to facilitate training of NL-to-FOL translation systems and enable their systematic comparison. In practice, however, only a limited number of partially human-curated resources are publicly available, most notably FOLIO \citep{DBLP:conf/emnlp/HanS0QRZCPQBSWS24} and MALLS \citep{DBLP:conf/acl/YangXPSF24}. Prior work has examined limitations in evaluation metrics and task design
\citep{brunello2026llms}, but the reference annotations themselves have received little systematic attention.

In this paper, we show that these annotations contain significant errors. We conduct a systematic human audit of the \textsc{FOLIO} validation split and a subset of \textsc{MALLS} test instances, finding that 42.5\% and 42\% of examples, respectively, contain incorrect FOL formalizations.
Additionally, 17.8\% of \textsc{FOLIO} and 51\% of \textsc{MALLS}
include NL statements that admit multiple, non-equivalent yet defensible interpretations (Section~\ref{sec:data-quality}). These annotation errors have a direct and notable impact on evaluation: re-assessing state-of-the-art models (Gemma~4 31B-it, Qwen3-30B-A3B, GPT-4o-mini) against our corrected ground truth yields accuracy gains of +11 to +23 percentage points 
(Section~\ref{sec:performance-shift}).

Exhaustive manual review, however, does not scale to large datasets or real-world deployments. We therefore introduce an LLM-assisted oversight framework that focuses human effort on the instances most likely to be incorrect.  The approach exploits a key empirical finding: LLMs rarely flag a correct formalization as wrong---in the \textsc{FOLIO} validation split, Gemma marks only ${\sim}3\%$ of correct instances as incorrect (see Appendix \ref{app:verdict_accuracy}). Directing annotators to instances that the model deems incorrect, our framework enables 90\% annotation accuracy while requiring review of only ${\sim}20\%$ of \textsc{FOLIO} validation instances and ${\sim}5\%$ of the \textsc{MALLS} test subset.
Moreover, when applied to the high-quality private dataset \textsc{GGC} \citep{barker2011student}, it introduces negligible additional noise, confirming its reliability when errors are rare.

Our framework finds natural application across several settings where high-quality FOL corpora are required but exhaustive human verification is expensive. 
First, in \emph{dataset curation}, rather than subjecting every instance to costly expert review, curators can focus exclusively on the small fraction flagged as suspicious, turning the process into a much more tractable task. 
Second, in \emph{formal methods} and verification, the framework can assist experts by identifying likely incorrect formalizations of NL specifications, acting as a warning layer.
Finally, in \emph{runtime verification} scenarios~\citep{DBLP:journals/corr/abs-2504-21022}, where newly defined behavioral constraints must be checked immediately against live data, there is no time for manual review; the framework provides on-the-fly feedback on whether a generated formula faithfully captures the intended requirement.

\paragraph{Contributions.}
\begin{enumerate}
    \item \textbf{Corrected datasets:} revised FOL formalizations and NLI labels (when applicable) for the \textsc{FOLIO} validation split (275 instances) and for a \textsc{MALLS} test subset (100 instances) (Section~\ref{sec:data-quality}).
    \item \textbf{Evaluation impact:} accuracy gains (from +11 to +23 percentage points) when state-of-the-art LLMs are assessed against the corrected ground truth (Section~\ref{sec:performance-shift}).
    \item \textbf{LLM-assisted oversight framework:} a novel
    strategy that directs human annotation effort toward instances most likely to contain errors, substantially reducing review cost without sacrificing quality of the revision (Section~\ref{sec:curation}).
\end{enumerate}
We release the corrected datasets and all code of our oversight
framework at \url{https://github.com/dslab-uniud/Fixing-FOLIO-and-MALLS.git}.

\section{Datasets and Their Quality Analysis}
\label{sec:data-quality}

We analyze errors and ambiguities in two datasets splits routinely used by the community for NL-to-FOL evaluation:  \textsc{FOLIO} validation split \citep{DBLP:conf/emnlp/HanS0QRZCPQBSWS24} and  a subset of the \textsc{MALLS} test set \citep{DBLP:conf/acl/YangXPSF24}. We describe in the following our methodology for human verification and correction. As shown in Table~\ref{tab:curation-summary}, approximately 42.5\% of \textsc{FOLIO} and 42\% of \textsc{MALLS} instances contain annotation errors, with direct consequences for model evaluation
(Section~\ref{sec:performance-shift}).

\begin{table*}[t]
\centering
\small
\begin{tabular}{lccc}
\hline
\textbf{Quantity} & \textbf{FOLIO} & \textbf{MALLS} & \textbf{GGC} \\
\hline
Reviewed Dataset Size & 275 & 100 & 213 \\
\hline
Incorrect FOL sentences (\# / \%) & \texttt{117/42.5\% } & \texttt{42/42\%} & \texttt{0/0.0\%} \\
 -- among which syntactically incorrect (\# / \%) & \texttt{29/10.5\%} & \texttt{3/3\%} & \texttt{0/0.0\%} \\
  -- among which semantically incorrect (\# / \%) & \texttt{88/32.0\%} & \texttt{39/39\%} & \texttt{0/0.0\%} \\
\hline
Ambiguous NL sentences (\# / \%) & \texttt{49/17.8\%} & \texttt{51/51\%} & \texttt{18/8.4\%} \\
\hline
Ambiguous NL and incorrect FOL (\# / \%) & \texttt{37/13.5\%} & \texttt{19/19\%} & \texttt{0/0.0\%} \\
\hline
\end{tabular}
\caption{Errors and ambiguities identified during human curation. \textit{Incorrect FOL sentences} counts formalizations that are wrong in the original datasets; subcategories are mutually exclusive.}
\label{tab:curation-summary}
\end{table*}

\subsection{Datasets}

\textbf{FOLIO} is an NLI-through-FOL benchmark: each instance contains a multi-sentence story (premises) and a conclusion, both in NL and FOL, paired with a NLI label (\textsc{True}/\textsc{False}/\textsc{Unknown}) indicating whether the conclusion is entailed by, contradicted by, or independent of the premises. The dataset is split randomly into training, validation, and test sets (${\sim}$1360/275/275 instances).\footnote{Errors found in the validation split may hence apply to the other
splits as well.} FOLIO is the dominant benchmark for NL-to-FOL
translation \citep{MALLS, brunello2026llms} and neurosymbolic reasoning \citep{DBLP:conf/emnlp/OlaussonGLZSTL23, DBLP:conf/iclr/RyuKLY25}.
We verify its validation split (\textsc{FOLIO\_validation}).

\textbf{MALLS} is a large-scale autoformalization dataset of NL--FOL pairs synthetically generated by GPT-4, commonly used for training autoformalizers  \citep{journals/corr/abs-2509-22338, DBLP:journals/corr/abs-2409-16461}. The full dataset comprises ${\sim}$28K instances; 1000 were declared human-checked and reserved as a test set.
We consider the first 100 of these (\textsc{MALLS\_test}).\footnote{The MALLS test set provides no metadata about the ordering of its instances, and the original paper does not document any explicit ordering criterion. In the absence of such information, no sampling strategy is demonstrably more representative than any other; we take the first 100 instances, considering it a transparent and reproducible choice.}

We also include 213 instances from \textbf{GGC} \citep{barker2011student} as a control, to verify that our curation framework does not introduce noise into already-correct annotations. GGC contains students' FOL
submissions for \emph{Tarski's World} exercises from \emph{Language, Proof and Logic} \citep{LPL}, validated by instructors via an automatic tool, providing strong guarantees of syntactic and semantic correctness.\footnote{GGC is not publicly available but is shared by the authors upon request.}

\subsection{Quality Analysis}
\label{sec:quality-analysis}
Prior works \citep{ DBLP:conf/emnlp/OlaussonGLZSTL23,brunello2026llms} have noted the presence of errors in FOLIO and MALLS; this paper aims to quantify their impact on evaluation outcomes.

In both datasets we identified both \textit{translation errors}  and \textit{ambiguous NL sentences}. Both phenomena may penalize a model that produces correct formalizations: translation errors cause a valid formula to be scored against an incorrect reference, while ambiguous sentences allow multiple legitimate interpretations, only one of which is encoded in the ground truth. GGC showed no significant errors, with only a small
portion of ambiguous sentences (${\sim}8.5\%$).

The original \textsc{FOLIO\_validation} also contains \textit{NLI labels inconsistent} with the provided formalizations; we corrected 17 such labels
(${\sim}8.4\%$ of the dataset) and re-verified them against NL reasoning.

In the following, we detail the annotation protocol, the issues identified during curation and the choices made to address them.

\paragraph{Annotation Protocol.}
The curation was conducted by two experienced reviewers---a PhD student and a research fellow, both with a mathematical background including coursework in logic. For each dataset, instances were split equally between the two annotators: each first reviewed their own half independently, and then cross-checked the half reviewed by the other. At this stage, the inter-annotator disagreement rate on the \textit{correctness} of the original dataset formulas (i.e., cases where one annotator judged a formalization correct and the other did not) was 9.1\% for FOLIO and 7\% for MALLS. The disagreement rate on \textit{ambiguity} (i.e., cases where one annotator judged a sentence ambiguous and the other did not) was 10.2\% for FOLIO and 13\% for MALLS.\footnote{Percentages are computed over the total number of instances processed.} All disagreements were subsequently resolved through discussion until consensus was reached.

\paragraph{Ontology Issues. }
Since no dataset provides an explicit ontology, we extract the signature---predicates, arities, and constants---automatically from the FOL formula and then we use an LLM to assign natural language meanings to each symbol, verifying then their consistency manually.

We encountered two distinct types of ontological issues. The first is \emph{inconsistencies}: when a symbol appeared with conflicting arities or meanings across the premises and conclusion of the same story, or within the same formula, we manually corrected the ontology to restore consistency. In \textsc{MALLS\_test}, we frequently found compound relational symbols (e.g., \texttt{WorksInNewsIndustryAndReportsOnEvents}) that obscure the logical structure. In such cases, we decomposed the predicate into its constituent parts (e.g., \texttt{WorksInNewsIndustry} and \texttt{ReportsOnEvents}) to recover a compositional ontology.

The second type is \emph{ontological looseness}: unconventional but internally consistent logical conventions applied uniformly throughout a formalization. A concrete example is the following: the NL expression \textit{school events} is encoded as the constant \texttt{schoolEvent}, even though FOL constants canonically denote specific individuals rather than classes. A more standard rendering would use a predicate \texttt{SchoolEvent(x)} ranging over individuals. This pattern recurs across several FOLIO stories. 

Unlike the first type, cases of ontological looseness do not compromise the validity of the reasoning. Since ontology design admits genuine degrees of freedom, there is no principled basis for preferring a finer or coarser granularity in the absence of additional contextual information; therefore, instances exhibiting these features are \emph{not} regarded as errors in the dataset. In some cases, for the sake of adhering to standard conventions, we nonetheless rename constants or predicates, or reify concepts; however, as already noted, this \lq\lq stylistic choice\rq\rq{} does not affect the error count reported in Table \ref{tab:curation-summary}.

\paragraph{Translation Errors. }
We identified recurring errors that compromise the correctness of the ground truth. \textit{Syntactic errors} (parenthesis mismatches, symbol typos, ontology misuse, and free variables) affect 10.5\% of instances in FOLIO and 3\% in MALLS. The remaining are \textit{semantic errors}, where a syntactically valid formula misrepresents the NL meaning: these are more prevalent ($\sim$32\% in FOLIO and $\sim$39\% in MALLS) and harder to detect automatically. They include quantifier scope errors, missing information explicitly stated in the NL sentence, incorrect entity relativization, and broader logical structure failures.

A summary of these findings is reported in Table~\ref{tab:curation-summary}. Appendix~\ref{app:FOLIO_errors_discussion} discusses how to reconcile these findings with seemingly contradictory LLM logical translation performance rates reported in the literature. The curated dataset releases include corrected ontologies (see Appendix~\ref{appendix:signature}), and corrected formalizations together with explanations of each correction (see Appendix~\ref{appendix:data-format-human}, \ref{appendix:FOLIOExample}, and \ref{appendix:MALLS} for details).

\paragraph{Ambiguities. }
\label{sec:ambiguity}
The inherent ambiguity of NL sentences poses a fundamental challenge for NL-to-FOL formalization \citep{DBLP:journals/aiopen/YadavPS21}: when acknowledged, it is typically addressed by requiring that sentences explicitly state all background information necessary for their interpretation, or that vague terms be avoided. Since no prior work provides systematic criteria for identifying ambiguity in NL-to-FOL benchmarks, we adopt an operative definition and annotate ambiguous instances explicitly during the human curation phase. We regard a sentence as ambiguous when, given a fixed ontology, multiple non-equivalent formalizations are defensible.

Ambiguity has direct consequences on evaluation: because all NL-to-FOL benchmarks supply a single reference formula, equivalence-based scoring  penalizes a model when it proposes correct alternative formalizations.

Explicit ambiguity annotation thus serves a dual purpose: it enables downstream analyses to account for this confound, and it fosters a cleaner assessment of model capabilities. For a qualitative classification of the ambiguity types encountered across datasets, see Appendix~\ref{appendix:ambiguity}.

\paragraph{FOLIO's NLI Labels Inconsistencies.}
\label{sec:FOLIO-labels}
As already discussed, FOLIO is also an NLI dataset. For each set of premises $\{p_1, \dots, p_n\}$ and conclusion $c$, there is also a label $l \in \{$\textsc{True}, \textsc{False}, \textsc{Unknown}$\}$ that reflects if $c$ is entailed by $p_1, \dots, p_n$ (\textsc{True}), is in contradiction with them (\textsc{False}) or its truth value can't be deductively established (\textsc{Unknown}). 

FOLIO provides formalizations $\varphi_1, \dots, \varphi_n$ of the story premises and $\psi$ of the conclusion. These can be used to check whether the label $l$ assigned to a pair premises-conclusion is correct: indeed, the entailment should be preserved in FOL, so the value of $l$ should depend on whether from $\varphi_1, \dots, \varphi_n$, one could derive $\psi$, or its negation, or neither of them. 

Using the original FOL sentences, the label provided by the author is different from the one derived from the Z3 automatic solver \citep{de2008z3} 18\% of times: this is largely motivated by the great amount of errors present in the FOL formulas, and partially (2\%) by the fact that the inference works only assuming some background knowledge that is not explicitated in the NL sentence and it is not known if it can be assumed (hence formalized) or not.

Our curation (i) corrects the NLI labels (8.4\% of the entire dataset), (ii) makes implicit background knowledge explicit when necessary for the entailment (through the \texttt{NLI\_requires\_more\_info} tag), (iii) re-verifies the consistence between the new labels and the curated formalizations with the Z3 solver.

\subsection{LLM Re-evaluation and Performance Shifts}
\label{sec:performance-shift}

\begin{table*}[t]
    \centering\small
    \resizebox{\linewidth}{!}{%
    \begin{tabular}{l | c c c | c c c | c c c | c c c}
        \multicolumn{13}{c}{\textbf{Translation Accuracy [\%]}} \\
        \hline
        & \multicolumn{6}{c|}{\textbf{FOLIO}} & \multicolumn{6}{c}{\textbf{MALLS}} \\
        \cline{2-13}
        \multirow{2}{*}{\textbf{Model}} 
            & \multicolumn{3}{c|}{\textbf{All Instances}} 
            & \multicolumn{3}{c|}{\textbf{Unambiguous Only}} 
            & \multicolumn{3}{c|}{\textbf{All Instances}} 
            & \multicolumn{3}{c}{\textbf{Unambiguous Only}} \\
        & Orig. & Cur. & $\Delta$ 
        & Orig. & Cur. & $\Delta$ 
        & Orig. & Cur. & $\Delta$ 
        & Orig. & Cur. & $\Delta$ \\
        \hline
        Gemma~4 31B-it  
            & 50.9 & \textbf{74.2} & \textbf{+23.3}
            & 59.7 & \textbf{82.7} & \textbf{+23.0}
            & 66.0 & \textbf{86.0} & \textbf{+20.0} 
            & 61.2 & \textbf{93.9} & \textbf{+32.7} \\
        Qwen3-30B-A3B   
            & 47.5 & 58.5 & \textbf{+11.0} 
            & 55.2 & 65.0 & \textbf{+9.8}
            & 52.1 & 68.0 & \textbf{+15.9} 
            & 53.3 & 65.3 & \textbf{+12.0} \\
        GPT-4o-mini     
            & 48.4 & 61.5 & \textbf{+13.1} 
            & 57.1 & 69.0 & \textbf{+11.9}
            & 49.0 & 68.0 & \textbf{+19.0}  
            & 40.8 & 65.3 & \textbf{+24.5} \\
        \hline
    \end{tabular}%
    }
    \caption{Translation accuracy (\%) on \textsc{FOLIO\_validation} and \textsc{MALLS\_test} under \emph{Original} (Orig.) and \emph{Curated} (Cur.) ground truth for all instances and the unambiguous subset. $\Delta$ = (Curated -- Original); all values are positive, confirming that curation consistently raises measured accuracy.}
    \label{tab:performance-shift}
\end{table*}

We re-evaluate a selection of recent LLMs---Gemma~4 31B-it~\cite{gemmateam2025gemma3technicalreport}, Qwen3-30B-A3B~\cite{yang2025qwen3technicalreport}, and GPT-4o-mini~\cite{openai2024gpt4ocard}---on the NL-to-FOL translation task using Chain-of-Thought prompting with the target ontology provided in context.

Models are evaluated on both the original and curated datasets: given a NL sentence $p$ and an ontology $\Omega$, we ask the model to formalize $p$ into a formula $\varphi$ adhering to the ontology. We check (binary pass/fail) whether $\varphi$ is logically equivalent to the reference formula: accuracy is the fraction of instances that are equivalent to the reference. In the original (uncurated) setting, the model receives the original ontology and is scored against the original dataset formula as reference; in the curated setting, it receives our curated ontology and is scored against our curated formula (see Appendix \ref{appendix:setup} for details). As shown in Table~\ref{tab:performance-shift}, curation consistently raises measured accuracy across all models, with gains of +11 to +23 percentage points on \textsc{FOLIO} and +16 to +20 on \textsc{MALLS}.
The largest improvements are exhibited by Gemma~4, the strongest model, suggesting that annotation noise disproportionately penalizes models capable of producing valid but non-reference formalizations. We also report results restricted to unambiguous instances, which offer a clear illustration of the impact of ambiguity on autoformalization datasets and provide a cleaner assessment of model capability that is not confounded by inherently ambiguous NL sentences: with respect to the total dataset, if we restrict on the unambiguous subset, Gemma~4 performs +8.5 and +7.9 percentage points better on FOLIO and MALLS, respectively.

\section{LLM-assisted Human Oversight}
\label{sec:curation}
Manual inspection is the most reliable way to ensure ground-truth quality, but does not scale: costs (money and time) grow linearly with dataset size, making exhaustive review impractical for large corpora. Delegating verification to automation is equally unsatisfactory, since assessing the alignment between a NL sentence and its formal representation remains an open problem \cite{fol-survey, DBLP:conf/aaai/MensfeltCFKTS26}. We therefore investigate a middle ground: using LLMs as judges \cite{DBLP:conf/nips/ZhengC00WZL0LXZ23} to build a first-pass filter that concentrates human attention on instances most likely to contain errors. To this end, in this section we describe the \emph{Verdict-and-Refinement Task}, two pipeline designs that instantiate it, and the resulting prioritization strategy for LLM-assisted dataset curation.

\subsection{Verdict-and-Refinement Task}
\label{sec:task-def}
We define the \emph{Verdict-and-Refinement} task (V\&R) as follows: given a NL sentence $p$, a candidate FOL formula $\varphi$, and an ontology $\Omega$ (predicates, constants, and their associated meanings), the goal is to produce (i) a \emph{verdict} $v \in \{\textsf{yes}, \textsf{no}, \textsf{?}\}$ and (ii) a \emph{refinement} FOL formula $\psi$. Specifically:
\begin{itemize}
\item $v = \textsf{yes}$: $\varphi$ correctly formalizes $p$ in the given ontology $\Omega$; no refinement is needed;
\item $v = \textsf{no}$: $\varphi$ incorrectly formalizes $p$ under $\Omega$; the refinement $\psi$ should provide a correct formalization of $p$;
\item $v = \textsf{?}$: the model is uncertain, or $p$ is ambiguous, i.e., it admits multiple non-equivalent yet defensible formalizations under $\Omega$: in the latter case, $\psi$ should capture the most natural interpretation of $p$.
\end{itemize}

The three-way verdict reflects a deliberate design choice, driven primarily by the treatment of ambiguity. A binary \textsf{yes}/\textsf{no} decision would force the model to either reject legitimate alternative interpretations or arbitrarily privilege one reading as canonical; the \textsf{?} label instead explicitly accommodates ambiguity while also allowing abstention under uncertainty.

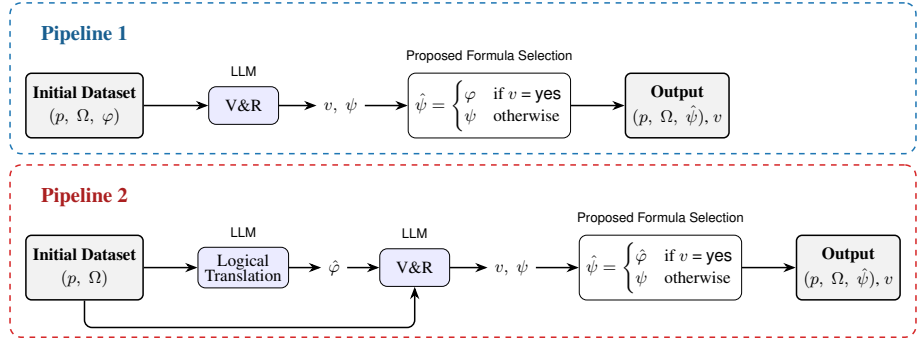
\begin{figure*}[h!]
\centering
\resizebox{0.75\linewidth}{!}{%
\begin{tikzpicture}[
    node distance=0.5cm and 0.7cm,
    box/.style={draw, rounded corners, minimum height=0.7cm, minimum width=1.3cm,
                align=center, font=\small},
    llmbox/.style={box, fill=blue!8},
    dataset/.style={draw, thick, rounded corners=3pt, fill=gray!10,
                    minimum height=1.2cm, minimum width=1.6cm,
                    align=center, font=\small},
    selbox/.style={draw, rounded corners, minimum height=1.0cm, minimum width=1.8cm,
                   align=center, font=\small},
    arr/.style={-Stealth, thick},
    llmlabel/.style={font=\scriptsize\sffamily, text=black!70!black},
    pipeline1box/.style={draw=tabblue, dashed, thick, rounded corners=6pt, inner sep=8pt},
    pipeline2box/.style={draw=tabred,  dashed, thick, rounded corners=6pt, inner sep=8pt},
    ]
  \node[font=\bfseries,, text=tabblue!80!black] (t1) {Pipeline 1};

  \node[dataset, below=0.4cm of t1] (ds1) {%
    {\bfseries Initial Dataset}\\[2pt]
    $(p,\;\Omega,\;\varphi)$};

  \node[llmbox, right=1.2cm of ds1] (judge1) {V\&R};
  \node[llmlabel, above=0.05cm of judge1] {LLM};

  \node[font=\small, right=0.7cm of judge1] (out1) {$v,\;\psi$};

  \node[selbox, right=0.8cm of out1] (sel1) {%
    $\hat{\psi}=\begin{cases}\varphi & \text{if $v$ = \textsf{yes}}\\ \psi & \text{otherwise}\end{cases}$};
  \node[llmlabel, above=0.05cm of sel1] {Proposed Formula Selection};

  \node[dataset, right=1.0cm of sel1] (cur1) {%
    {\bfseries Output}\\[2pt]
    $(p,\;\Omega,\;\hat{\psi})$, $v$};

  \draw[arr] (ds1) -- (judge1);
  \draw[arr] (judge1) -- (out1);
  \draw[arr] (out1) -- (sel1);
  \draw[arr] (sel1) -- (cur1);

  \node[font=\bfseries, text=tabred!80!black, below=2.4cm of t1] (t2) {Pipeline 2};

  \node[dataset, below=0.4cm of t2] (ds2) {%
    {\bfseries Initial Dataset}\\[2pt]
    $(p,\;\Omega)$};

  \node[llmbox, right=1.0cm of ds2] (gen) {Logical \\[-2pt]Translation};
  \node[llmlabel, above=0.05cm of gen] {LLM};

  \node[font=\small, right=0.6cm of gen] (phihat) {$\hat{\varphi}$};

  \node[llmbox, right=0.6cm of phihat] (judge2) {V\&R};
  \node[llmlabel, above=0.05cm of judge2] {LLM};

  \node[font=\small, right=0.7cm of judge2] (out2) {$v,\;\psi$};

  \node[selbox, right=0.8cm of out2] (sel2) {%
    $\hat{\psi}=\begin{cases}\hat\varphi & \text{if $v$ = \textsf{yes}}\\ \psi & \text{otherwise}\end{cases}$};
  \node[llmlabel, above=0.05cm of sel2] {Proposed Formula Selection};

  \node[dataset, right=1.0cm of sel2] (cur2) {%
    {\bfseries Output}\\[2pt]
    $(p,\;\Omega,\;\hat{\psi})$, $v$};

  \draw[arr] (ds2) -- (gen);
  \draw[arr] (gen) -- (phihat);
  \draw[arr] (phihat) -- (judge2);
  \draw[arr] (judge2) -- (out2);
  \draw[arr] (out2) -- (sel2);
  \draw[arr] (sel2) -- (cur2);

  \draw[arr, rounded corners=4pt] (ds2.south) -- ++(0,-0.5) -| (judge2.south);

    \coordinate (p1ext) at (cur2.east |- cur1.north);
    \coordinate (p2ext) at (cur2.south |- ds2.south);
    \coordinate (p2extB) at ($(p2ext) + (0, -0.4cm)$);  
    
    \begin{pgfonlayer}{background}
      \node[pipeline1box, fit=(t1)(ds1)(judge1)(out1)(sel1)(cur1)(p1ext)] {};
      \node[pipeline2box, fit=(t2)(ds2)(gen)(phihat)(judge2)(out2)(sel2)(cur2)(p2extB)] {};
    \end{pgfonlayer}
    
\end{tikzpicture}%
}
\caption{The two pipelines. Each starts from the \textit{Initial Dataset}
containing triplets $(p, \Omega,  \varphi)$, and produces an output with the \textit{Formalization Proposal} $(p, \Omega,\hat\psi)$ and the \textit{verdict} $v$.
\textit{Pipeline~1} judges the original 
formula~$\varphi$ directly.
\textit{Pipeline~2} first re-generates a candidate~$\hat{\varphi}$
from~$p$ and $\Omega$ alone, then judges it.}
\label{fig:pipelines}
\end{figure*}

\subsection{Pipelines}
\label{sec:pipelines} 

Considering a dataset whose instances are triplets $(p, \Omega, \varphi)$ as in Section~\ref{sec:task-def}, we embed the V\&R task within two distinct pipelines, illustrated in Figure~\ref{fig:pipelines}.

\begin{itemize}
  \item \textbf{Pipeline 1: Direct V\&R.} The LLM receives the triplet $(p, \Omega, \varphi)$ and produces a verdict $v$ and, optionally, a refinement $\psi$ when $\varphi$ is judged incorrect or ambiguous. The ultimately proposed formula is $\hat\psi = \varphi$ if $v = \textsf{yes}$, and
  $\hat\psi = \psi$ otherwise.
  \item \textbf{Pipeline 2: Re-generation and V\&R.} The original formula $\varphi$ is discarded. The LLM first translates $p$ under $\Omega$ into a candidate $\hat\varphi$, which is then submitted to the same V\&R step as in Pipeline~1. The finally proposed formula is $\hat\psi = \hat\varphi$ if $v = \textsf{yes}$, and $\hat\psi = \psi$ otherwise.
\end{itemize}

The two pipelines embody different assumptions about the original formula $\varphi$.
Pipeline~1 treats it as a reliable anchor: refining an existing
formula may be more effective than generating one from scratch.
Pipeline~2 treats it as a potential liability: when $\varphi$ is
unreliable or difficult for the model to evaluate, retaining it
may mislead the V\&R step, and a freshly generated candidate can be a safer starting point.

Each pipeline processes the original dataset and produces an output comprising two elements for each instance: (i) the \textit{formalization proposal} consisting of triplets $(p, \Omega, \hat\psi)$, where $\hat\psi$ is the pipeline's proposed formula, and (ii) the \textit{verdict} $v$ emitted during the V\&R step, referring to the original formula $\varphi$ or to the LLM-generated candidate $\hat\varphi$, respectively in Pipeline 1 and Pipeline 2.
Despite this difference in reference, in both cases the verdict carries information about the reliability of $\hat\psi$: a \textsf{yes} verdict signals that $\hat\psi$ derives from a formula that was deemed correct; \textsf{no} indicates it refines a formula deemed incorrect; and \textsf{?} flags either ambiguity in $p$ or model uncertainty. 

\subsection{Prioritization Order}
\label{sec:prioritize}
Assume a pipeline is run on all instances of a dataset, producing outputs $(p, \Omega, \hat\psi)$, each paired with a verdict $v$.
The verdicts $v$ can be used to prioritize manual inspection, by inspecting first the instances in a certain verdict class, followed by others from another class, and so on: we call this a \emph{prioritization order}. Intuitively, for example, a \textsf{yes} verdict means formula $\hat{\psi}$ was considered correct by both the generator (Pipeline 1: dataset creator; Pipeline 2: LLM) and the V\&R judge, making it the least likely to contain errors, and hence the last to review. 
In Section~\ref{sec:results}, we show how to find an adequate prioritization order.

\section{Experimental Setup}
\label{sec:setup}

To evaluate our proposed LLM-assisted oversight approach, we require ground-truth annotations for each instance. We therefore assume every original triplet $(p, \Omega, \varphi)$ is accompanied by (i) a gold verdict $v_{\text{gold}} \in \{\textsf{yes}, \textsf{no}, \textsf{?}\}$ assessing the correctness of $\varphi$ or ambiguity of $p$, and (ii) a gold formalization $\varphi_{\text{gold}}$ of $p$ under $\Omega$.

Specifically, for each instance $(p, \varphi)$ in the original \textsc{FOLIO\_validation} and \textsc{MALLS\_test}, we derive the ontology $\Omega$ from $\varphi$ and construct the curated ground-truth $\varphi_\text{gold}$; given this, we derive the verdict $v_\text{gold}$ marking as \textsf{yes}, \textsf{no}, and \textsf{?} the instances for which $\varphi \equiv \varphi_\text{gold}$, for which  $\varphi \not \equiv \varphi_\text{gold}$, and for which $p$ is ambiguous. This is exactly the process used in Section~\ref{sec:curation} for curating the datasets. We also perform our evaluation on \textsc{GGC} to measure the possible degradation introduced by the pipelines when applied to already-correct data. Since GGC annotations are verified correct, $v_{\text{gold}} = \textsf{yes}$ and $\varphi_{\text{gold}} = \varphi$ for all instances, except for the few ambiguous sentences (where $v_{\text{gold}} = \textsf{?}$).

\paragraph{Simulating human review.}
Our primary interest is in measuring how effectively each proposed pipeline, when coupled with a prioritization order, concentrates errors at the top of the inspection queue, so that a human annotator achieves high accuracy while reviewing as few instances as possible. We model this by simulating a perfect annotator who, upon inspecting an instance, immediately corrects it. We track \emph{accuracy}---the fraction of correct instances in the dataset---as a function of the fraction of instances reviewed by the human, $h \in [0,1]$; we refer to this as the
\emph{accuracy--human effort curve}.

Two baselines provide reference curves against which to compare the prioritization effectiveness of the pipelines:
(i) \textbf{Black Baseline (original dataset)}---the annotator reviews instances from the original dataset in random order. Accuracy starts at the original dataset accuracy and increases linearly to 100\% as $h$ reaches 1;
 (ii) \textbf{Green Baseline (LLM regeneration)}---the annotator starts from a synthetic dataset produced in the Logical Translation phase of Pipeline~2 and reviews instances in random order. Accuracy starts at the regenerated dataset accuracy and increases linearly to 100\%  as $h$ reaches 1.

\paragraph{Metrics.}
As we mentioned, each strategy corresponds to an accuracy--human effort curve, with
steeper early rises indicating more effective prioritization. Our
primary metric is the \emph{Area Under the Curve} (AUC; higher is
better). Since AUC alone is not easily interpretable, we complement
it with $T_\alpha$, the minimum fraction of instances to review to reach $\alpha\%$ accuracy (lower is better). 
We report $T_{90}$ and $T_{95}$.

\paragraph{Models and prompting.} We run all pipelines with three models: GPT-4o-mini, Qwen3-30B-A3B, and Gemma~4 31B-it. Following best practices for LLM evaluation \cite{mizrahi-etal-2024-state, DBLP:conf/iclr/Sclar0TS24}, and since our goal is not to identify a single optimal prompt but rather to assess whether the capabilities required for our application can be consistently elicited, for each model we considered four prompting strategies varying in reasoning approach (standard CoT (B1), few-shot (B2), few-shot + CoT (B3), and meta-prompting (B4)) and four prompt variants (pv1–pv4) differing in their degree of skepticism toward the input formula and level of instructional detail, yielding a total of 16 prompting combinations.
Full details are in Appendix~\ref{appendix:setup}.

\section{Results and Discussion}
\label{sec:results}

\begin{figure*}[t]
    \centering
    \includegraphics[width = 0.8 \linewidth, trim = {0 1cm 0 3cm}, clip]{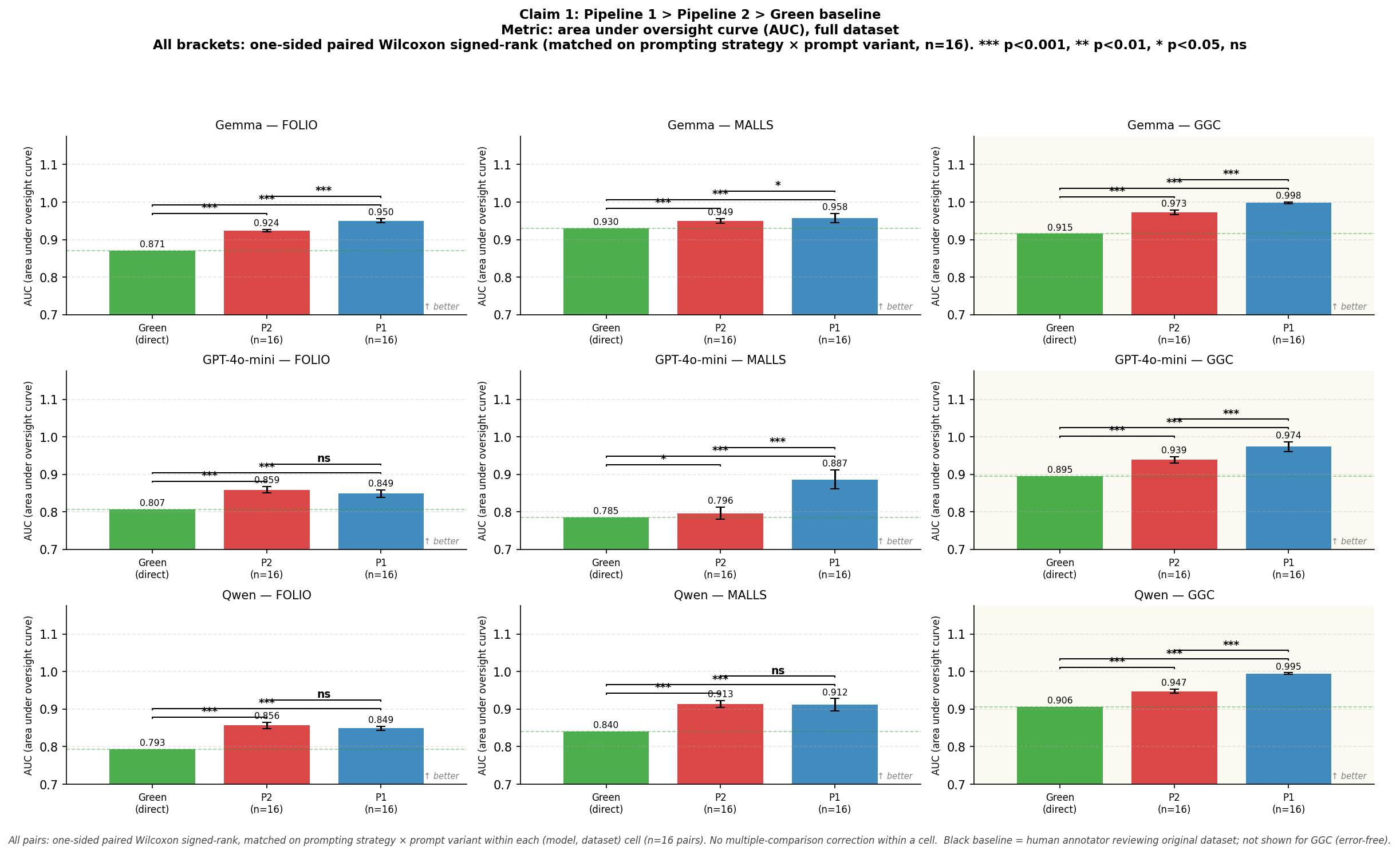}
    \caption{Pipelines comparison across models (horizontally) and datasets (vertically) according to the AUC (Area Under the Curve) metric. Asterisks denote statistical significance (*** = $p < 0.001$, ** = $p < 0.01$, * = $p < 0.05$) under the one-sided paired Wilcoxon signed-rank test. \textsf{ns}: non-significant.}
    \label{fig:claim1}
\end{figure*} 
\begin{figure*}[t]
    \centering
    \includegraphics[width = 0.8\linewidth, trim = {0 1cm 0 3.5cm}, clip]{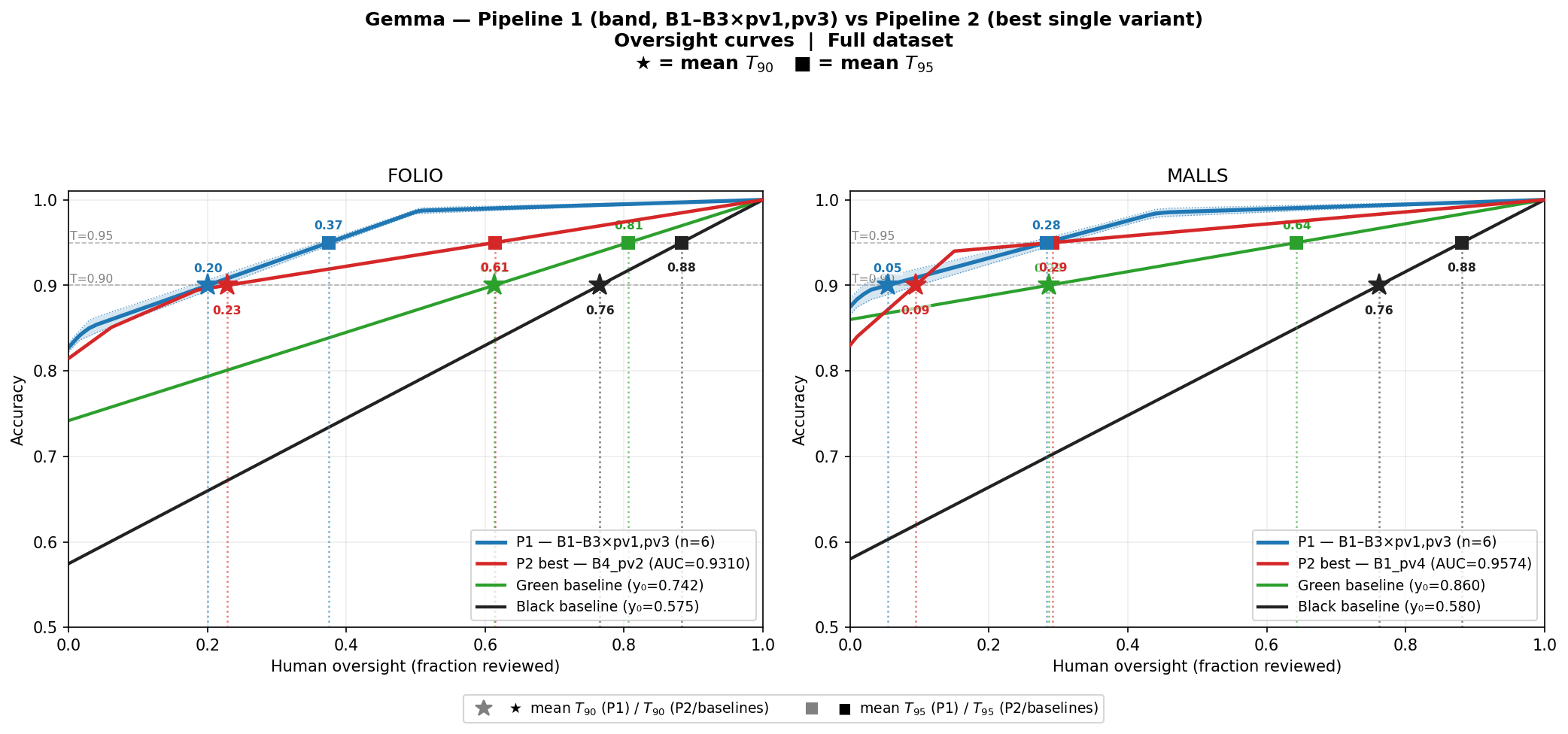}
    \caption{\emph{Accuracy-human effort} curve. Each plot shows Gemma's performance on  \textsc{FOLIO\_validation} (left) and \textsc{MALLS\_test} (right). The blue band represents Pipeline 1 (min/max and average across the six prompting combinations \{B1, B2, B3\}$\times$\{pv1, pv3\}); the red curve shows the best-AUC configuration of Pipeline~2; the black and green line represents respectively the Black and Green Baseline.}
    \label{fig:claim3}
\end{figure*}

For each of the two pipelines, in Appendix \ref{app:verdict_accuracy}, we measure accuracy within each verdict class and find that it increases monotonically from \textsf{?} to \textsf{no} to \textsf{yes}. 
We thus adopt the following prioritization order for both of them: (i)~malformed outputs, (ii)~\textsf{?}, (iii)~\textsf{no}, (iv)~\textsf{yes}. All results below are reported under this prioritization.\footnote{Note that other prioritization strategies can, in principle, be adopted, including strategies that do not rely on the verdict produced by the V\&R task. See Appendix \ref{app:prioritazion_strategy} for an alternative proposal of this kind.}

Figure~\ref{fig:claim1} compares the prioritization strategies across datasets and models. In terms of AUC, Pipeline~1 (blue) outperforms Pipeline~2 (red), which in turn outperforms the Green Baseline (green). This ordering holds across all metrics ($T_{90}$, $T_{95}$) and is statistically significant (Appendix~\ref{app:pipeline_comparison}).\footnote{For simplicity, these figures omit the Black Baseline, which is based on the original dataset; its (bad) performance is addressed below.}
Among the models evaluated, Gemma performs best (Appendix~\ref{app:model_comparison}). 

Given the witnessed results, we focus on Gemma with Pipeline~1 for the main analysis.
Among the prompting strategies, B4 and variants pv2 and pv4 perform
significantly worse than the remaining configurations; the others
(B1, B2, B3 with pv1 and pv3) are statistically indistinguishable,
indicating that Pipeline~1 is robust to prompting choices within this
set (Appendix~\ref{app:variant_selection}). See Appendix \ref{app:full-results} for the results for all models and datasets.

Figure~\ref{fig:claim3} plots the \emph{accuracy--human effort curves}, showing the full range of Pipeline~1
performance across the retained prompting configurations (blue band)
and the best-AUC configuration for Pipeline~2. The ordering among
Green Baseline, Pipeline~2, and Pipeline~1 is preserved across all
datasets and metrics.

The steep initial slopes confirm that the above prioritization order effectively front-loads error-prone instances. The practical gains are substantial: on \textsc{FOLIO}, 90\% accuracy is reached after reviewing only 20\% of instances--- a third of the effort required by the Green Baseline (61\%) and less than a third of that required by the Black Baseline (76\%). On \textsc{MALLS}, the reduction is even more pronounced: 5\% suffices against 29\% for the Green Baseline (a $>5\times$ reduction) and against 76\% for the Black Baseline (a $15\times$ reduction). Comparable gains hold for $T_{95}$.

Furthermore, in both datasets, instances assigned a \textsf{yes} verdict
constitute around half of the data; accuracy restricted to such a class is above 97\%. This opens the possibility of treating such a class as a
ready-to-use curated training set; however, doing so halves the
dataset and may disproportionately discard the most challenging
instances---a trade-off that practitioners should weigh carefully.

Finally, when applied to the already 100\% correct dataset GGC, all three models with Pipeline 1 produce a limited degradation of $\leq 5\%$ (see Appendix \ref{app:full-results}).

\section{Conclusions and Future Works}
\label{sec:limitations}
This work addresses systematic annotation errors in two widely used
NL-to-FOL datasets, \textsc{FOLIO\_validation} and \textsc{MALLS\_test},
uncovering error rates of 42.5\% and 42\% respectively. These errors
measurably distort LLM evaluation, with performance estimates shifting
by up to 23 points once corrected annotations are used.
To address the infeasibility of exhaustive human oversight, we propose
an LLM-assisted framework that concentrates human effort on the
instances most likely to contain errors. The best pipeline
configurations allow a human annotator to reach 90\% accuracy after
reviewing only 20\% of \textsc{FOLIO\_validation} and 5\% of
\textsc{MALLS\_test}.

Future investigation includes: (i) as a natural extension, applying the pipeline
to the full training splits of \textsc{FOLIO} and \textsc{MALLS},
where exhaustive manual review is prohibitive but systematic quality
improvements would have broad impact; (ii) increasing pipeline
autonomy through iterative refinement loops and automated ontology
construction would reduce the remaining manual effort, (iii)
introducing a dedicated pre-processing stage to detect and resolve
linguistic ambiguity prior to formalization is a promising direction
toward a more robust curation process.

\section{Limitations}
\label{sec:conclusion}
Dataset curation is inherently delicate and error-prone. Despite the strategies adopted to remain as objective as possible, curation remains an inherently subjective process, and residual annotation mistakes may persist. We therefore expect that the exact error and ambiguity counts reported here could shift slightly upon further scrutiny. However, this does not undermine our core message: these datasets contain a substantial number of annotation errors and ambiguities that meaningfully compromise the reliability of LLM evaluation on them. 
Annotators corrected instances, without access to the outputs of the models evaluated in
Section~\ref{sec:performance-shift} and Section~\ref{sec:results},
ruling out any risk of bias toward the reasoning patterns of a
specific tested model. 
A further limitation in this regard concerns ambiguity: the criteria used to flag
ambiguous sentences are not standardized by prior work, and alternative
strategies may be equally defensible. 

Finally, the proposed prioritization strategy is shown to work well for the models and datasets tested, and it relies on the empirical evidence that LLMs rarely flag correct formalizations as incorrect. However, when applying the strategy to a new dataset or a new model, we advise validating this underlying assumption in the new setting as well, using a small calibration set.

\newpage
\bibliography{references}

\appendix

\section{Autoformalization in FOL: Background}
\label{appendix:task-def}

\paragraph{First-Order Logic.}
A \emph{signature} $\sigma = (\V, \C, \F, \R)$ consists of a countably infinite set of
variables $\V$, a set of constants $\C$, a set of function symbols $\F$, and
a set of relation/predicate symbols $\R$ (each with an associated arity), where all four sets
are pairwise disjoint.
\emph{Terms} are built inductively: variables and constants are base terms; if
$t_1,\dots,t_n$ are terms and $f \in \F$ has arity $n$, then $f(t_1,\dots,t_n)$ is a
term.
\emph{FOL formulas} over $\sigma$ are defined inductively by:
\begin{align*}
  \varphi \ := \ &r(t_1,\dots,t_n) \ | \ \neg\varphi \ | \
  \varphi \land \varphi \ | \ \varphi \lor \varphi \ | \ \varphi \oplus \varphi \ | \\
  & \varphi \to \varphi \ | \ \varphi \leftrightarrow \varphi \ | \
  \exists x\,\varphi \ | \ \forall x\,\varphi,
\end{align*}
where $r \in \R$ with arity $n$, the $t_i$ are terms, and $\neg, \land, \lor, \oplus, \to, \leftrightarrow,
\exists, \forall$ denote negation, conjunction, inclusive disjunction, exclusive disjunction, implication, equivalence,
and the existential and universal quantifiers respectively.

\paragraph{Operator precedence.}
When parentheses are omitted, the logical operators above are assumed to bind in the
following order (from highest to lowest precedence):
\begin{equation*}
  \{\neg,\,\exists,\,\forall\} \;\succ\; \land \;\succ\; \lor
  \;\succ\; \to \;\succ\; \leftrightarrow\; \succ \oplus.
\end{equation*}
For example, $\neg A \land B \to C \lor D$ is parsed as
$((\neg A) \land B) \to (C \lor D)$.
Parentheses are required explicitly whenever the intended scoping departs from this convention.

\paragraph{Ontology.}
Evaluating whether a FOL formula faithfully represents a natural language sentence
requires an interpretation of the symbols in the formula. We use the term
\emph{ontology} to refer to the signature enriched with the intended NL
meaning of each symbol (e.g., $\text{Tet}(x)$ means ``$x$ is a tetrahedron''). We denote this usually with the symbol $\Omega$.

\paragraph{The autoformalization task.}
Autoformalization is the task of producing a FOL representation that symbolically
captures the semantic meaning of a NL phrase $p$. Following~\cite{brunello2026llms}, the task
decomposes along two dimensions:
\begin{itemize}
  \item \textbf{Ontology Extraction (OE):} identifying a suitable signature 
  for the formalization, along with the interpretation of the symbols in it.
  \item \textbf{Logical Translation (LT):} translating $p$ into a FOL formula using
  the symbols and meanings provided by a given ontology.
\end{itemize}

This decomposition arises from the fact that determining whether a given formula is equivalent to another is only possible when the two formulas share the same signature. Therefore, in this paper, whenever we ask a model to produce a formalization, we also provide the signature as part of the input (LT task). This allows us to reliably evaluate performance on the formalization task automatically.

As noted in \cite{brunello2026llms}, without a labeled signature it is also impossible to manually assess whether a formula conveys the same meaning as a natural language sentence, since the interpretation of the symbols used in the formula would remain unclear.

For a more complete definition of the FOL-autoformalization task and for a survey of approaches, datasets, and open chellenges in NL-to-FOL formalization, see \cite{fol-survey}.

\section{Dataset Error Rates are Coherent with Recent Literature}
\label{app:FOLIO_errors_discussion}

At first glance, the high percentage of incorrect formalizations we identify in the \textsc{FOLIO\_validation} dataset (42.5\%) may seem to contradict recent findings in the literature. For instance, \citep{brunello2026llms} demonstrated that state-of-the-art models like o3-mini can achieve approximately 80\% accuracy in NL to FOL translation tasks on the train split of FOLIO dataset.

\paragraph{Sentence-Level vs. Mixed-Instance Evaluation. }
The discrepancy stems from the fundamental unit of analysis. Evaluations that report $\sim$80\% accuracy assess \emph{isolated sentences} translated into standalone FOL formulas. In contrast, our quality assessment of FOLIO evaluates the dataset according to its actual NLI structure. 

The \textsc{FOLIO\_validation} split contains 275 total instances, divided into two distinct types:
\begin{enumerate}
    \item \textbf{Stories (73 instances):} The concatenated premises of an NLI problem, averaging roughly 6 natural language sentences each.
    \item \textbf{Conclusions (202 instances):} Single-sentence logical conclusions.
\end{enumerate}
In the context of automated reasoning, a story's formalization is only correct if \emph{every single component sentence} is flawlessly translated and ontologically consistent. The reason behind the choice to consider the story as a whole is that some information is not explicit in specific sentences and derives from the context posed by the premises in the story: thus, if we have in mind general applications of the autoformalization task,  ours seems the most natural choice.

\paragraph{Mathematical Justification. }
We can demonstrate that our 42.5\% instance-level error rate is entirely consistent with $\sim$80\% sentence-level accuracy as reported in the literature.

The \textsc{FOLIO\_validation} dataset contains 275 instances composed of 73 multi-sentence stories (averaging 6 sentences each) and 202 single-sentence conclusions, yielding approximately $73 \times 6 + 202 = 640$ total sentences.
Supposing that any error done by the model in FOLIO stories are located in only one sentence, based on our error rate, we get $ 42.5\%\times 275 \sim 117$ incorrect sentences. A perfect annotator then would have produced an accuracy of:
\[
\frac{640 - 117}{640} \sim 82\%
\]
which is consistent with the $\sim$80\% accuracy reported by \citep{brunello2026llms} when evaluating state-of-the-art models on isolated FOLIO sentences.

We also notice that the dataset used for the evaluation in \cite{brunello2026llms} is different from the one used here (\textsc{FOLIO\_training} vs. \textsc{FOLIO\_validation}) even if they are random splits coming from the same pool.

\section{Instance of The Ontology File}
\label{appendix:signature}

We report here an example of the entries of the ontology file given within the code.
The following JSON is what can be found for the FOLIO instance \texttt{story\_380}. Each entry in
\texttt{Rel} corresponds to a predicate symbol extracted from the curated FOL formula, annotated with its arity (\texttt{arity}) and NL meanings (i.e., its NL interpretation (\texttt{pos\_meaning}), and the meaning when the predicate is negated (\texttt{neg\_meaning})); argument positions are indicated by \texttt{\{0\}}, \texttt{\{1\}}, etc. Each entry in \texttt{Const} maps a logical constant name to its NL interpretation as it appears in the following example.

For FOLIO, each ontology is described by the \texttt{story\_id}: for each premises in the multi-sentence story and for each conclusion related to the story, the formalization in FOL provided in the curated dataset adhere to the symbols here listes. For MALLS, each sentence comes with an own ontology: hence, we will have 73 entries for FOLIO and 100 for MALLS.

\begin{jsonblock}
{
  "story_id": 380,
  "Rel": {
    "InThisClub":     { "arity": 1, "pos_meaning": "{0} is in this club",
                                    "neg_meaning": "{0} is not in this club" },
    "PerformOftenIn": { "arity": 2, "pos_meaning": "{0} performs often in {1}",
                                    "neg_meaning": "{0} does not perform often in {1}" },
    "Attend":         { "arity": 2, "pos_meaning": "{0} attends {1}",
                                    "neg_meaning": "{0} does not attend {1}" },
    "VeryEngagedWith":{ "arity": 2, "pos_meaning": "{0} is very engaged with {1}",
                                    "neg_meaning": "{0} is not very engaged with {1}" },
    "InActive":       { "arity": 1, "pos_meaning": "{0} is inactive",
                                    "neg_meaning": "{0} is not inactive" },
    "Disinterested":  { "arity": 1, "pos_meaning": "{0} is disinterested",
                                    "neg_meaning": "{0} is not disinterested" },
    "MemberOf":       { "arity": 2, "pos_meaning": "{0} is a member of {1}",
                                    "neg_meaning": "{0} is not a member of {1}" },
    "Chaperone":      { "arity": 2, "pos_meaning": "{0} chaperones {1}",
                                    "neg_meaning": "{0} does not chaperone {1}" },
    "Student":        { "arity": 1, "pos_meaning": "{0} is a student",
                                    "neg_meaning": "{0} is not a student" },
    "AttendSchool":   { "arity": 1, "pos_meaning": "{0} attends school",
                                    "neg_meaning": "{0} does not attend school" },
    "YoungChild":     { "arity": 1, "pos_meaning": "{0} is a young child",
                                    "neg_meaning": "{0} is not a young child" },
    "Teenager":       { "arity": 1, "pos_meaning": "{0} is a teenager",
                                    "neg_meaning": "{0} is not a teenager" },
    "WishToFurther":  { "arity": 2, "pos_meaning": "{0} wishes to further {1}",
                                    "neg_meaning": "{0} does not wish to further {1}" }
  },
  "Const": {
    "bonnie":                "Bonnie",
    "schoolTalentShow":      "the school talent show",
    "schoolEvent":           "school events",
    "community":             "the community",
    "highSchoolDance":       "high school dances",
    "academicCareer":        "academic career and educational opportunities",
    "educationalOpportunity":"educational opportunities"
  }
}
\end{jsonblock}

\section{Released Data Format Of Human-Annotated Subsets}
\label{appendix:data-format-human}
We release the human-annotated portions of FOLIO (275 instances) and MALLS (100 instances) as JSON records designed for reliable evaluations.

\paragraph{FOLIO: story records and conclusion records.}
We store each FOLIO instance as a \emph{story record} (containing the full multi-sentence NL premises (story-level context) together with its FOL translation), or as a \emph{conclusion record} (each record contains a conclusion in NL with its FOL counterpart). An example is reported in Appendix \ref{appendix:FOLIOExample}.

\paragraph{Common fields.}
Both record types share:
\begin{itemize}
    \item \texttt{id}: unique identifier (e.g., \texttt{story\_471}, \texttt{concl\_471\_20}).\footnote{For the story records, the last number identifies the story ID in the original dataset. For the conclusion records, the first number identifies the story ID to which the conclusion is linked, while the second number identifies the row in which the conclusion appears in the original version of the dataset hosted on Hugging Face, available at \url{https://huggingface.co/datasets/yale-nlp/FOLIO/viewer/default/validation}.}
    \item \texttt{NL\_sentence}: the NL text.
    \item \texttt{FOL\_sentence\_old}: original formula from the source dataset.
    \item \texttt{FOL\_sentence\_new}: curated formula (equal to \texttt{FOL\_sentence\_old} if unchanged).
    \item \texttt{corrected}: whether the new FOL formula (\texttt{FOL\_sentence\_new}) is different from the original one (\texttt{FOL\_sentence\_old}). 
    \item \texttt{is\_originally\_correct}: whether the original FOL formula (\texttt{FOL\_sentence\_old}) is considered correct or incorrect, when computing the errors of the dataset. Notice that there can be cases in which \texttt{corrected} is \'yes\' and also \texttt{is\_originally\_correct} also: in this case, it means that we are in the case of \emph{ontological loseness} discussed in Section \ref{sec:curation} and some usually adopted conventions (about namings or concepts reification) are applied to the original formulas.
\end{itemize}

\paragraph{Ambiguity metadata.}
If the NL admits multiple plausible readings, we set:
\begin{itemize}
    \item \texttt{ambiguity}: boolean.
    \item \texttt{ambiguity\_explanation}: categorical tag \texttt{some\_plural}, \texttt{either\_or}, \texttt{option\_partition}, \texttt{sufficient\_necessary\_condition}, \texttt{determinative\_the}, \texttt{existential\_universal}  when possible or otherwise an explanation of the ambiguity we detected.
\end{itemize}
Even when \texttt{ambiguity=true} we provide, in the \texttt{FOL\_sentence} field, the most plausible alternative given the instructions detailed in Section Section~\ref{sec:ambiguity}.

\paragraph{Correction explanation.}
When \texttt{corrected=true}, we include:
\begin{itemize}
    \item \texttt{correction\_explanation}: short description of what was changed and why.
\end{itemize}

\paragraph{NLI labels (conclusion records only).}
Conclusion records additionally include:
\begin{itemize}
    \item \texttt{label\_old}: the original FOLIO NLI label.
    \item \texttt{label\_old\_Z3}: the NLI label derivable from the original formalizations
    \item \texttt{label}: our corrected label (coherent with \texttt{label\_Z3})
    \item \texttt{label\_Z3}: the NLI label derivable from the new formalizations (plus eventually the information in the \texttt{NLI\_requires\_more\_info}.
    \item \texttt{NLI\_requires\_more\_info}: the FOL formula that represents the background knowledge that must be added to the story to ensure that the NLI label is consistent with the FOL entailment.
\end{itemize}

\paragraph{MALLS: item records.}
MALLS uses a single-item format, so each instance corresponds to a unique sentence always. Each record includes \texttt{id},
\texttt{NL\_sentence}, \texttt{FOL\_sentence\_old}, \texttt{FOL\_sentence\_new},
\texttt{corrected}, and (when applicable) \texttt{ambiguity} with
\texttt{ambiguity\_explanation} as explained above. You can find an example in the Appendix \ref{appendix:MALLS}.

\paragraph{Ontology.}
Each FOLIO story with conclusions and MALLS instance is further accompanied by an \emph{ontology} (signature enriched with LLM-generated glosses): see Appendix~\ref{appendix:signature}.

\section{Example of Released \textsc{FOLIO\_validation\_curated} Dataset}
\label{appendix:FOLIOExample}

\noindent\textbf{Story record.}
\begin{jsonblock}
{
 {"id": "story_380", 
 "NL_sentence": "People in this club who perform in school talent shows often attend and are very engaged with school events. People in this club either perform in school talent shows often or are inactive and disinterested community members. People in this club who chaperone high school dances are not students who attend the school. All people in this club who are inactive and disinterested members of their community chaperone high school dances. All young children and teenagers in this club who wish to further their academic careers and educational opportunities are students who attend the school. Bonnie is in this club and she either both attends and is very engaged with school events and is a student who attends the school or is not someone who both attends and is very engaged with school events and is not a student who attends the school.", 
 "FOL_sentence_new": "∀x (InThisClub(x) ∧ PerformOftenIn(x, schoolTalentShow) → Attend(x, schoolEvent) ∧ VeryEngagedWith(x, schoolEvent)) ∧ ∀x (InThisClub(x) → (PerformOftenIn(x, schoolTalentShow) ⊕ (InActive(x) ∧ Disinterested(x) ∧ MemberOf(x, community)))) ∧ ∀x (InThisClub(x) ∧ Chaperone(x, highSchoolDance) → ¬(Student(x) ∧ AttendSchool(x))) ∧ ∀x (InThisClub(x) ∧ (InActive(x) ∧ Disinterested(x) ∧ MemberOf(x, community)) → Chaperone(x, highSchoolDance)) ∧ ∀x (InThisClub(x) ∧ (YoungChild(x) ∨ Teenager(x)) ∧ WishToFurther(x, academicCareer) ∧ WishToFurther(x, educationalOpportunity) → (Student(x) ∧ AttendSchool(x))) ∧ InThisClub(bonnie) ∧ ((Attend(bonnie, schoolEvent) ∧ VeryEngagedWith(bonnie, schoolEvent) ∧ Student(bonnie) ∧ AttendSchool(bonnie)) ⊕ (¬(Attend(bonnie, schoolEvent) ∧ VeryEngagedWith(bonnie, schoolEvent)) ∧ ¬(Student(bonnie) ∧ AttendSchool(bonnie))))",
 "corrected": "yes", 
 "ambiguity": true, 
 "ambiguity_explanation": "either_or", 
 "correction_explanation": "1. Spelling: Corrected 'Studen' to 'Student' and 'bonne' to 'bonnie'. 2. Linkage: 'highSchoolDances' changed to 'highSchoolDance' in Premise 4. 3. Missing Parenthesis & Operators: Fixed the missing opening parenthesis for the implication in Premise 5. Changed 'YoungChildren' to 'YoungChild' and XOR '⊕' to inclusive OR '∨' in Premise 5 to match 'young children and teenagers'. Added missing 'educationalOpportunity' to WishToFurther to fully capture the NL. 4. Premise 6: Removed hallucinated outer parentheses added by the previous reviewer and properly bounded the variables to 'bonnie'.", 
 "FOL_sentence_old": "∀x (InThisClub(x) ∧ PerformOftenIn(x, schoolTalentShow) → Attend(x, schoolEvent) ∧ VeryEngagedWith(x, schoolEvent))\n∀x (InThisClub(x) → PerformOftenIn(x, schoolTalentShow) ⊕ (InActive(x) ∧ Disinterested(x) ∧ MemberOf(x, community)))\n∀x (InThisClub(x) ∧ Chaperone(x, highSchoolDance) → ¬(Studen(x) ∧ AttendSchool(x)))\n∀x (InThisClub(x) ∧ (InActive(x) ∧ Disinterested(x) ∧ MemberOf(x, community)) → Chaperone(x, highSchoolDances))\n∀x (InThisClub(x) ∧ (YoungChildren(x) ⊕ Teenager(x)) ∧ WishToFurther(x, academicCareer)) → Studen(x) ∧ AttendSchool(x))\nInThisClub(bonnie) ∧ ¬((Attend(x, schoolEvent) ∧ VeryEngagedWith(bonnie, schoolEvent)) ⊕ (Studen(bonne) ∧ AttendSchool(bonnie)))", 
 "is_original_correct": "no"}
}
\end{jsonblock}

\noindent\textbf{Conclusion record (example).}
\begin{jsonblock}
{
  "id": "concl_380_1",
  "NL_sentence": "Bonnie performs in school talent shows often.",
  "FOL_sentence": "PerformOftenIn(bonnie, schoolTalentShow)",
  "label": "Uncertain",
  "corrected": "yes",
  "correction_explanation": "1. Predicate Match: Corrected 'Perform' to 'PerformOftenIn' to match the premises. 2. Removed Extraneous Conjunction: The original included 'InThisClub(bonnie)' which is not stated in the conclusion NL.",
  "FOL_sentence_old": "InThisClub(bonnie) ∧ (Perform(bonnie, schoolTalentShow))",
  "label_old": "Uncertain"
}
\end{jsonblock}

\section{Example of Released \textsc{MALLS\_test\_curated} Dataset Example}
\label{appendix:MALLS}

\begin{jsonblock}
{
  "NL_sentence": "A vacation is relaxing if it includes beautiful scenery and enjoyable activities.",
  "FOL_sentence_old": "∀x (Vacation(x) ∧ Relaxing(x) → (BeautifulScenery(x) ∧ EnjoyableActivities(x)))",
  "FOL_sentence_new": "∀x (Vacation(x) → ((BeautifulScenery(x) ∧ EnjoyableActivities(x)) → Relaxing(x)))",
  "corrected": "yes",
  "correction_explanation": "1. The original formula inverted the implication symbols.",
  "id": 1
}
\end{jsonblock}

\section{Classification of Common Ambiguities}
\label{appendix:ambiguity}

We report a classification of the recurring NL ambiguities we detected during
human oversight of FOLIO and MALLS. Each type is described with an explanation,
a concrete example showing the two plausible FOL readings, and the canonicalization
convention we adopt. This is not intended as a comprehensive taxonomy of NL ambiguity
in general; it is an empirically grounded catalogue of the patterns that arise
frequently in existing NL--FOL datasets for the autoformalization task.

For consistency, in the dataset curation procedure we adopt one interpretation for each category of ambiguity. Nevertheless, since these instances are flagged as ambiguous, users who wish to work with the dataset may choose to resolve ambiguous sentences differently from the choices we adopted here.

\paragraph{\texttt{some\_plural}.}
The word \textit{some} followed by a plural noun is ambiguous between asserting the
existence of at least one individual satisfying a property and asserting the existence
of at least two distinct individuals. For example, \textit{some pets are not mammals}
can be formalized as
$\exists x\, (\text{Pet}(x) \land \lnot \text{Mammal}(x))$
or as
$\exists x\, \exists y\, (\text{Pet}(x) \land \lnot \text{Mammal}(x) \land
\text{Pet}(y) \land \lnot \text{Mammal}(y) \land x \neq y)$.
We always adopt the one-witness reading (first formula) and flag the instance as ambiguous. We decide to flag this form of ambiguity since the original FOLIO dataset sometimes use the one-witness formulation and others the two-witnesses one.

\paragraph{\texttt{either\_or}.}
The connective \textit{either\ldots or} is ambiguous between inclusive disjunction
($\lor$) and exclusive disjunction ($\oplus$). For example,
\textit{Peter's pets are all either monkeys or birds} can be read inclusively as
$\forall x\, (\text{PetOfPeter}(x) \rightarrow (\text{Monkey}(x) \lor \text{Bird}(x)))$
or exclusively as
$\forall x\, (\text{PetOfPeter}(x) \rightarrow (\text{Monkey}(x) \oplus \text{Bird}(x)))$,
where the latter rules out any pet being both a monkey and a bird.
We always adopt the exclusive reading ($\oplus$) and flag the instance as ambiguous.

\paragraph{\texttt{option\_partition}.}
Enumerated categorization (e.g., \textit{$X$ can be categorized as $A$, $B$, and $C$})
is ambiguous between an inclusive partition (allowing overlap) and an exclusive
partition (enforcing disjointness). For example,
\textit{machine learning algorithms can be categorized as supervised learning,
unsupervised learning, and reinforcement learning} can be read inclusively as
$\forall x\, (\text{MLA}(x) \leftrightarrow (\text{SL}(x) \lor \text{UL}(x) \lor \text{RL}(x)))$,
or exclusively by additionally enforcing pairwise disjointness.
We flag the instance as ambiguous and adopt the inclusive reading ($\lor$), except when it is explicitly stated that the options are incompatible.

\paragraph{\texttt{sufficient\_necessary\_condition}.}
Conditional phrasing such as \textit{$X$ if $P$} can denote a sufficient condition
($P \rightarrow X$) or be intended as a biconditional ($X \leftrightarrow P$) when the intended meaning of the NL sentence is to \emph{define} a specific category of entities. For
example, \textit{An organization is non-profit if it has a charitable mission, does
not distribute profits to owners, and primarily relies on donations} can be
interpreted as
$\forall x\, ((\text{Organization}(x) \land \text{CharitableMission}(x) \land
\lnot \text{DistributesProfits}(x) \land \text{ReliesOnDonations}(x)) \rightarrow
\text{NonProfit}(x))$,
but a plausible alternative is a biconditional that also constrains \emph{all}
non-profits to satisfy those properties. We flag such cases as ambiguous and, when
not clearly disambiguated by the context, adopt the $P \rightarrow X$ interpretation.

\paragraph{\texttt{determinative\_the}.}
The definite article \textit{the} presupposes a unique, contextually salient referent,
but in isolation it is often unclear whether the formalization should encode uniqueness
explicitly. For example, \textit{the cube is red} could be formalized as a simple
existential $\exists x\, (\text{Cube}(x) \land \text{Red}(x))$, or with an explicit
uniqueness constraint
$\exists x\, (\text{Cube}(x) \land \text{Red}(x) \land \forall y\,
(\text{Cube}(y) \rightarrow y = x))$,
or even as a direct predication over a named constant when the referent has been
introduced earlier in a story. Without prior discourse context, the intended reading
is underspecified. We flag such cases as ambiguous and, when a single formalization is
required, default to the simple existential unless story-level context motivates the
uniqueness reading.

\paragraph{\texttt{existential\_universal}.}
Bare singular noun phrases of the form \textit{a $C$ has property $P$} are ambiguous
between a universal reading (every $C$ has $P$) and an existential one (some
particular $C$ has $P$). For example, \textit{a child has a mother} is most naturally
read as a universal generalization
$\forall x\, (\text{Child}(x) \rightarrow \exists y\, (\text{Mother}(y,x)))$,
yet without context, the same syntactic pattern could refer to a specific child. We
flag such cases as ambiguous and adopt the universal reading when the sentence
expresses a general rule, falling back to the existential reading when the sentence
clearly introduces a particular individual.

\section{Experimental Setup}
\label{appendix:setup}
This section specifies the experimental configuration for evaluating
both pipelines. We benchmark three LLMs using four prompting strategies
(B1--B4) and four prompt variants (pv1--pv4). All combinations are
evaluated on \textsc{FOLIO\_validation}, \textsc{MALLS\_test}, and
\textsc{GGC}. We describe the prompting strategies (Section~\ref{appendix:prompts-strategies}), the prompt
variants (Section~\ref{appendix:prompts-variants}),
the models (Section~\ref{sec:models}), the data splits (Section~\ref{sec:splits}),
and how the logical equivalence check is performed (Section~\ref{sec:oracle}). All experimental artifacts are released
in the accompanying code repository.

\subsection{Prompting Strategies (B1--B4)}
\label{appendix:prompts-strategies}
We benchmark LLMs using four prompting strategies. As defined in Section~\ref{sec:task-def}, the model receives $p$, $\varphi$, and $\Omega$, and must output a verdict $v$ alongside a refinement $\psi$.

\begin{description}
    \item[B1: Zero-Shot Chain-of-Thought.] \citep{wei2023chainofthought}.
    Task description and CoT instruction only; no in-context examples. The model
    reasons step-by-step over $(p, \varphi, \Omega)$ before producing a verdict and refinement. Establishes
    a floor for instruction-following and self-directed reasoning alone.
    \item[B2: Few-Shot (no CoT).] \citep{llmfewshot}.
    Task description plus 5 labeled examples covering all  verdict classes,
    each with its ontology. No reasoning chain is requested. Tests whether exposure
    to demonstrations alone suffices.
    \item[B3: Few-Shot + CoT.]
    Combines B2's in-context examples with B1's CoT
    instruction.
    \item[B4: Feature-Based Decomposition.]
    A form of metaprompting \cite{zhang2024metapromptingaisystems}. Rather than a
    holistic judgement, the model evaluates $\varphi$ against a checklist of four
    sub-properties---\emph{syntactic well-formedness}, \emph{ontological faithfulness},
    \emph{truth-conditional equivalence}, and \emph{ambiguity resolution}---before
    aggregating. Tests whether decomposition improves precision
    on subtle cases.
\end{description}

\begin{table*}[!h]
    \centering\small
    \begin{tabular}{l p{10cm}}
        \toprule
        \textbf{Strategy} & \textbf{Description} \\
        \midrule
        \textbf{B1: Zero-Shot CoT} & Task description and step-by-step reasoning instruction only \cite{wei2023chainofthought}. \\
        \textbf{B2: Few-Shot} & Task description plus labeled examples covering all verdict classes \cite{llmfewshot}. \\
        \textbf{B3: Few-Shot + CoT} & Combines B2's in-context demonstrations with B1's reasoning request. \\
        \textbf{B4: Decomposition} & Metaprompting checklist \cite{zhang2024metapromptingaisystems} focused on syntax, ontology coherence, equivalence, and ambiguity. \\
        \bottomrule
    \end{tabular}
    \caption{Overview of the four benchmarked prompting strategies.}
    \label{tab:baselines}
\end{table*}

\subsection{Prompt Variants (pv1--pv4)}
\label{appendix:prompts-variants}

For each prompting strategy B1--B4, we render four prompt variants that vary along two dimensions: the evaluator's stance toward the candidate formula and the specificity of evaluation guidance. 

\begin{description}
    \item[\textbf{pv1}] -- \emph{neutral, minimal.} A direct task definition
    positioning the model as an expert linguist-logician. No explicit
    guidance on skepticism or evaluation protocol; the model is expected
    to assess the formula on its own merits.
    \item[\textbf{pv2}] -- \emph{cautious, specified.} Introduces explicit skepticism,
    instructing the model to ``avoid false confidence'' and to default to `\texttt{?}'
    when uncertain. Any doubt about expressiveness, interpretation,
    or logical equivalence should trigger the uncertain verdict rather than a binary
    yes/no judgment.
    \item[\textbf{pv3}] -- \emph{neutral, detailed.} Similar to pv1 in stance,
but instructs the model to produce a thorough analysis of the
strengths and weaknesses of the candidate formalization before
committing to a verdict, encouraging deeper reasoning prior to
the final judgment.
    \item[\textbf{pv4}] -- \emph{adversarial, rigorous.}
    Adopts the most
    skeptical stance: the model is instructed to assume $\varphi$ is
    incorrect until an adversarial analysis fails to find any flaw. The
    model systematically probes for common error types (wrong
    quantifier, scope inversion, predicate mismatch) and assigns
    \textsf{yes} only after such a search yields no flaw.
\end{description}

All four variants maintain identical task definitions and examples
(where applicable) across the three models, ensuring that cross-model
comparisons isolate model capability from prompt engineering. We show the first prompt version (pv1) for the B1 prompting strategy below. The full
prompt texts are released with the code: the different blocks that builds the prompt combinations discussed above are available at the script \url{/core/strategies/prompts/common.py}; the examples used in the few-shot version at \url{/core/strategies/prompts/few_shot_examples.py}.

\begin{tcolorbox}[
  breakable,
  colback=gray!4,
  colframe=black!60,
  boxrule=0.5pt,
  arc=1mm,
  left=4pt, right=4pt, top=4pt, bottom=4pt,
  fonttitle=\bfseries\small,
  title=Prompt template (B1-pv1),
  width=\columnwidth,
  fontupper=\small
]
You are an expert linguist and logician evaluating whether a candidate First-Order Logic (FOL) formula $\hat\varphi$ correctly and completely formalizes a natural language sentence $p$ using only the symbols of a given ontology.

\smallskip
Assign a quality verdict $q \in \{\text{`yes'}, \text{`?'}, \text{`no'}\}$:
\begin{itemize}\setlength\itemsep{2pt}
  \item[\textbf{yes}] $\hat\varphi$ correctly and completely captures the semantic meaning of $p$ within the provided ontology as a valid FOL formula. Note that the chosen formalism and ontology restrictions limit expressiveness: there will be cases in which not all the information in the NL sentence can be retained when converting it into a FOL formula using the given symbols. In such cases, the formalization should nonetheless be considered correct if $\hat\varphi$ is the best possible approximation of $p$ in FOL given the provided ontology.
  \item[\textbf{no}] $\hat\varphi$ is incorrect or incomplete due to syntax errors, semantic hallucination, ontology mismatch, or logical flaws. The only symbols permitted are the logical ones ($\forall\ \exists\ \neg\ \land\ \lor\ \to\ \leftrightarrow\ \oplus\ =\ \neq\ \top\ \bot$), and those explicitly defined in the provided ontology.
  \item[\textbf{?}] A confident and definitive verdict cannot be assigned because the NL sentence admits multiple non-equivalent interpretations, or because there is uncertainty about its logical correctness, or for other reasons. In this case, your comment MUST begin with exactly one of the following tags: [AMBIGUITY], [UNCERTAINTY], or [OTHER].
\end{itemize}

\smallskip
When $q$ is ``no'' or ``?'', also produce a corrected formalization $\psi$ that addresses the identified issues. $\psi$ should be a rigorous, improved version of $\hat\varphi$ that resolves the specific problems you found. Ensure $\psi$ is a complete well-formed FOL formula, not a partial suggestion; when $p$ is ambiguous, $\psi$ should represent the other possible interpretation of $p$. If $q$ is ``yes'', put ``null'' for $\psi$.
 
 \smallskip
 Before producing the verdict, reason step by step.

\end{tcolorbox}

\subsection{Models}
\label{sec:models}

We benchmark three LLMs spanning a closed-weights commercial system, an
open-weights mixture-of-experts model, and an open-weights dense model:
\begin{description}
    \item[\textbf{GPT-4o-mini}] OpenAI's small frontier model, accessed
    through the OpenAI API (version \textsf{gpt-4o-mini-2024-07-18}). Training on API inputs was disabled to prevent the model from being
    updated with data from the private GGC dataset.
    \item[\textbf{Qwen3-30B-A3B}.] Alibaba's 30B-parameter mixture-of-experts model with 3B active parameters per token, run locally. Extended chain-of-thought reasoning (thinking mode) was
    enabled.
    \item[\textbf{Gemma~4 31B-it}.] Google's 31B-parameter instruction-tuned dense model, run locally. Extended
    chain-of-thought reasoning (thinking mode) was enabled.
\end{description}

For GPT-4o-mini and Qwen3-30B-A3B, the maximum output token limit was set to 2,500, with retries capped at 8,000 tokens. For Gemma 4, the limit was increased to 8,000 tokens -- and retries to 13,000 -- to accomodate the longer reasoning traces produced during the thinking
phase.

We performed the local experiments on a Dell PowerEdge R750 running Red Hat Enterprise Linux 8.7 (Ootpa). The system is equipped with two Intel(R) Xeon(R) Platinum 8360Y CPUs at 2.40 GHz (72 physical cores in total), 512 GB of RAM, 4 TB of disk storage, and an NVIDIA A100 GPU with 80 GB of VRAM.

\subsection{Data Splits}
\label{sec:splits}

Both pipelines are evaluated on the three curated datasets:
FOLIO\_\textsc{validation} (275 instances), MALLS\_\textsc{test} (100 instances), and the GGC dataset (213 instances).

\subsection{Logical Equivalence and SMT solver}
\label{sec:oracle}
All logical equivalence checks between FOL formulas are performed using
the Z3 SMT solver \citep{de2008z3}, with a 30-second wall-clock timeout
per query. This configuration is used uniformly for the re-evaluation
of LLM performance on the curated datasets, the pipelines evaluation,
and NLI label verification. Queries that exceed the timeout are mapped
to an \texttt{SolverTimeoutError} and excluded from metric computation.

Z3 is invoked under two configurations depending on the dataset:
\begin{description}
    \item[Raw (FOLIO and MALLS).] No additional axioms; equivalence is decided as a
    pure first-order question over the formulas themselves.
    \item[With axioms (GGC).] Since GGC formalizes objects in
    Tarski's World, the relational symbols used are governed by domain
    constraints. 41 axioms of the domain
    (symmetry of converses such as \textsf{Smaller}/\textsf{Larger}, the
    symmetry and reflexivity of \textsf{SameSize}, \textsf{SameShape},
    \textsf{SameCol}, \textsf{SameRow}, the mutual exclusivity of shape and
    size classes, and statements like
    $(\textsf{Larger}(x,y) \vee \textsf{Larger}(y,x)) \leftrightarrow
    \neg \textsf{SameSize}(x,y)$) are added as background theory. This axiom set is released with the code.
\end{description} 

Timeout errors arise exclusively on GGC, likely due to the added
axioms increasing solver complexity; they affect only 0.4\% of GGC
instances. Notice that timeout errors have no effect for the results discussed in Section~\ref{sec:performance-shift}.

\section{Justification of the Human-Review Prioritisation Order}
\label{app:verdict_accuracy}

\begin{figure*}[t]
  \centering
  \includegraphics[width=\linewidth, trim={0 0 0 4cm}, clip]{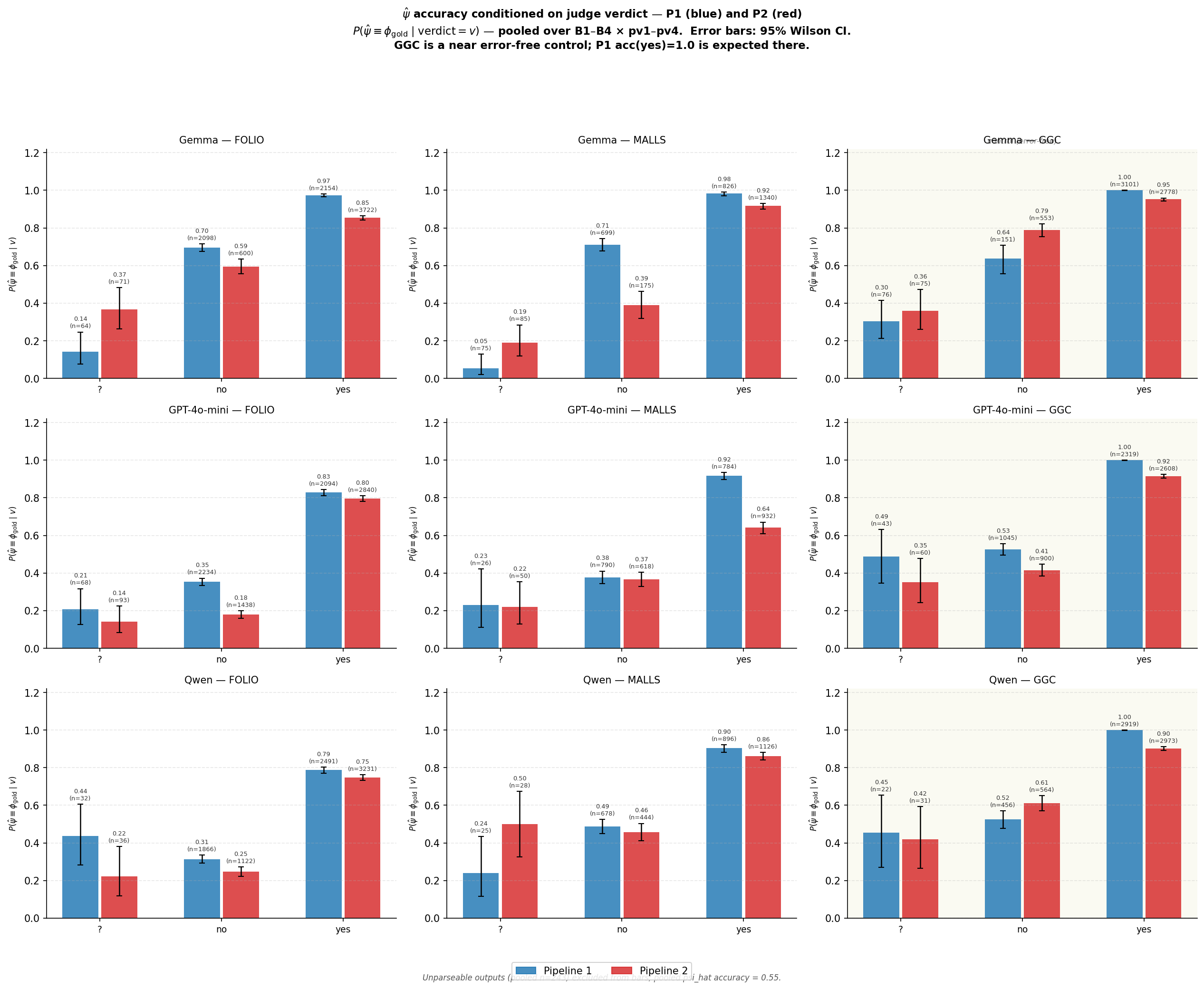}
  \caption{%
    Accuracy conditioned
    on the verdict assigned by the judge, for Pipeline~1 (blue) and
    Pipeline~2 (red).
    Each bar pools all
  eligible variants 
    (B1--B4~$\times$~pv1--pv4 for P1; self-judge B1--B4 for P2) within each
    cell,  yielding large
  counts and narrow confidence intervals
    Error bars are 95\% Wilson confidence intervals.
    \textsc{GGC} cells are shaded: because the original dataset is
    effectively error-free, every Pipeline 1 instance labelled \emph{yes} by the judge
    corresponds to a correct formula, making the conditioned accuracy equal to $1.00$.
    Unparseable outputs (total pooled $n=243$) are excluded from the bars
    because individual cells contain too few observations for reliable estimates.%
  }
  \label{fig:appF_verdict_accuracy}
\end{figure*}

\subsection{Motivation}
Each pipeline does not itself constitute a curation strategy; however,
the verdict $v$ carries information that can be exploited to define a
principled inspection order, directing human reviewers toward the
instances most likely to require correction. The rationale rests on a
key empirical claim: that accuracy varies systematically with the
verdict produced by the V\&R stage. Figure~\ref{fig:appF_verdict_accuracy}
provides the evidence for this claim.

\subsection{Accuracy by Verdict Group}

\paragraph{yes.}
Instances assigned verdict \emph{yes} achieve the highest $\hat\psi$ accuracy
in every model--dataset--pipeline cell.
Averaged over all 9 cells, Pipeline~1 reaches $0.933$ and Pipeline~2
reaches $0.843$.
Despite the gap between Pipeline 1 and Pipeline 2, even Pipeline~2's \emph{yes} group is substantially more
reliable than the other verdict groups, validating the policy of deferring
those instances to the last tier of human review.

\paragraph{no.}
Averaged over all 9 cells, Pipeline~1 reaches $0.514$ and Pipeline~2
reaches $0.45$.
These low values reflect two failure modes: (i)~the judge correctly flags
an erroneous formula and proposes a refinement that is itself incorrect, and
(ii)~the judge incorrectly flags a correct formula as \emph{no} and proposes
an unnecessary (and potentially wrong) refinement.
Both cases result in $\hat\psi \not\equiv \varphi_{\mathrm{gold}}$ and therefore
require human review, confirming that \emph{no}-verdict instances should be
prioritised.

\paragraph{?.}
Averaged over all 9 cells, Pipeline~1 reaches $0.283$ and Pipeline~2
reaches $0.308$. Crucially, \textsf{?} accuracy is lower than
\textsf{no} accuracy in most of the cells, empirically establishing
that \textsf{?} instances should be reviewed before \textsf{no}
instances in the queue.

\paragraph{Unparseable outputs.}
A small number of model outputs cannot be assigned to any verdict category
because they do not match the expected structured format.
Across all 9 cells the total count of such \emph{unparseable} responses is
243.
Pooling all cells, the $\hat\psi$ accuracy for unparseable outputs is
$131/243 \approx 0.54$.

This figure overstates
reliability: when the judge produces an unparseable output, the
pipeline falls back to the original formula ($\varphi$ in Pipeline 1, and $\hat\varphi$ in Pipeline 2) as $\hat\psi$
(Section~\ref{sec:pipelines}), so this accuracy merely reflects how often
the original formula happened to be correct. Unparseable outputs provide no information about whether
correction is needed and should be flagged for immediate review
regardless of apparent accuracy.

\paragraph{Summary of Prioritisation Order}
Combining the observations above, the pipeline's review order is:
\begin{enumerate}
  \item \textbf{Unparseable}: outcome unknown; review
        unconditionally.
  \item \textbf{?}: lowest accuracy; high
  probability of error or genuine ambiguity requiring human adjudication.
  \item \textbf{\textsf{no}:} moderate accuracy; likely
  erroneous but partially addressed by the proposed refinement.
  \item \textbf{\textsf{yes}:} highest accuracy; the
  judge is confident and usually correct; review only if capacity
  allows.
\end{enumerate}

This ordering directly determines the shape of the human-oversight curve:
the steepest accuracy gains per reviewed fraction occur in the first
tiers, so the area under the curve (AUC) rewards pipelines that concentrate
errors early in the queue.

\section{Other Prioritization Strategies}
\label{app:prioritazion_strategy}

In the paper we presented a prioritization strategy based on the verdict of the V\&R task; one may wonder whether other, more naive strategies could work just as well or better.

The first idea that comes to mind is to order dataset instances from the most difficult to the easiest: presumably, mistakes concentrate more on difficult instances than on easy ones. In the context of formalization it is not clear what ``difficulty'' means, but as a proxy ordering one can use the length of the NL sentence or of the FOL formula. To make this metric more robust to predicate/constant renaming, we count the number of \emph{words} for the NL sentence, and the number of \emph{logical elements} for the FOL formula, i.e., quantifiers (together with the variable they bind), logical operators, predicates, variables, and constants.\footnote{For example, under this notion of length the formula $\forall x (Human(x) \to Mortal(x))$ has length 6: $\forall x$, $Human$, $x$, $\to$, $Mortal$, $x$.}

This length measure induces a prioritization strategy when interpreted as a proxy for instance complexity: a human reviewer could curate the dataset starting from the longest instances and proceeding toward the shortest ones. This strategy can be applied either directly to the original (uncurated) dataset, or after an LLM has already produced a formalization of each NL sentence, in which case the reviewer orders instances by the length of the LLM-generated formula rather than the original one.

\begin{figure*}
    \centering
    \includegraphics[width = \linewidth, trim={0 0 0 2cm},clip]{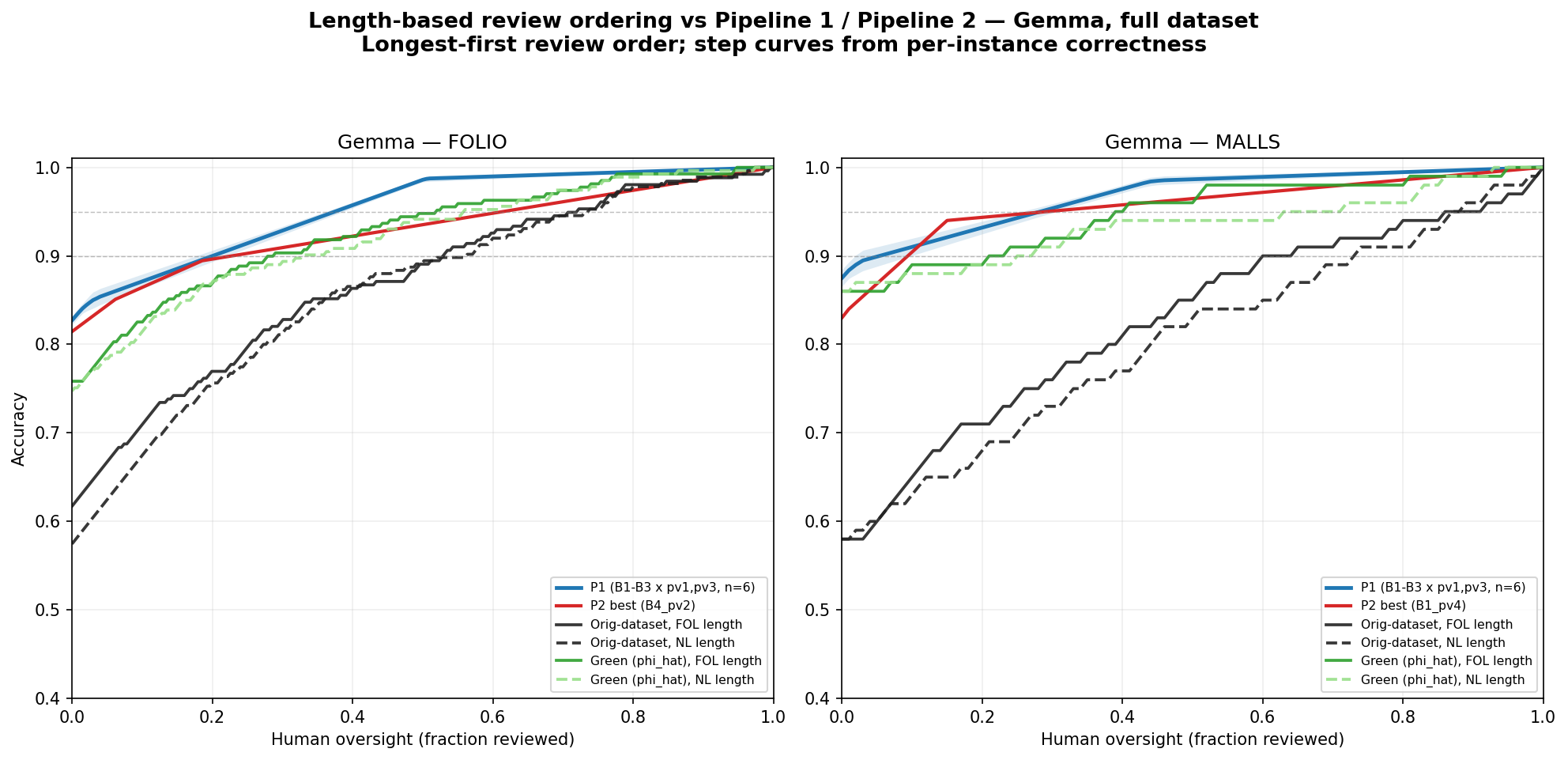}
    \caption{Comparison between Pipeline 1 (blue), Pipeline 2 (red), FOL and NL length based approach starting from the original dataset (black) and from the LLM (Gemma) formalizations (green).}
    \label{fig:length_ordering_comparison}
\end{figure*}

Figure~\ref{fig:length_ordering_comparison} compares these two length-based strategies (using both NL length and FOL length) against the V\&R prioritization strategy applied with Pipeline 1 and Pipeline 2 (Gemma), for FOLIO and MALLS. Using the convention used throughout the paper, in black we show the case in which we start from the original dataset, and in green when we start from the LLM (Gemma in this case) translations.

The figure shows that, despite these measures seem to correlate with the presence of annotation errors (the slope of the curves is bigger at the beginning and then decreases) our proposed V\&R approach outperforms these length-based strategies by a wide margin.

\section{Pipeline Comparison: $T_{90}$, $T_{95}$, and AAG}
\label{app:pipeline_comparison}

\begin{figure*}
    \centering
    \includegraphics[width = \linewidth, trim={0 1cm 0 3cm},clip]{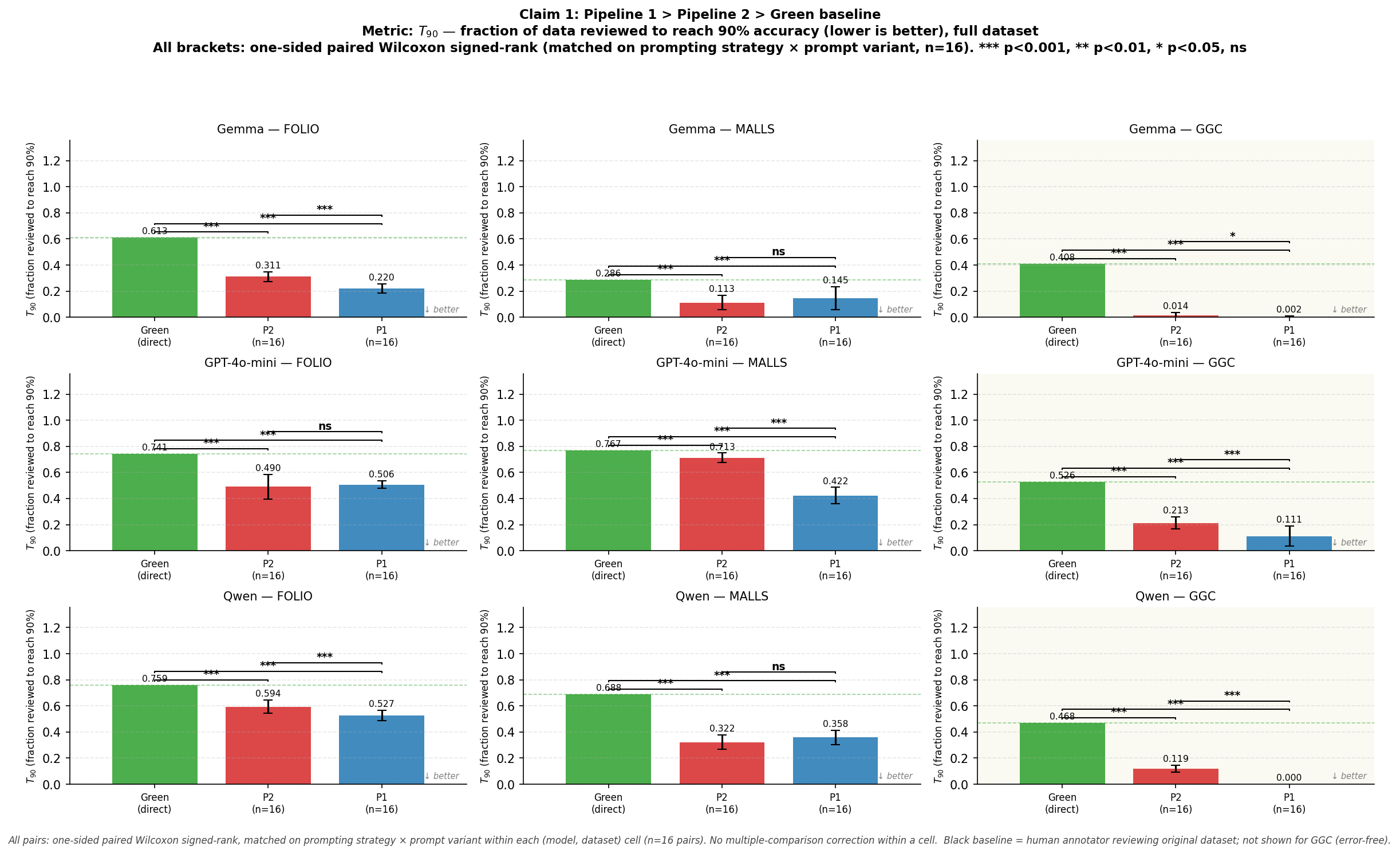}
    \caption{Pipeline comparison across models and datasets under the $T_{90}$ metric.}
    \label{fig:t90_pipeline_comparison}
\end{figure*}

\begin{figure*}
    \centering
    \includegraphics[width = \linewidth, trim={0 1cm 0 3cm},clip]{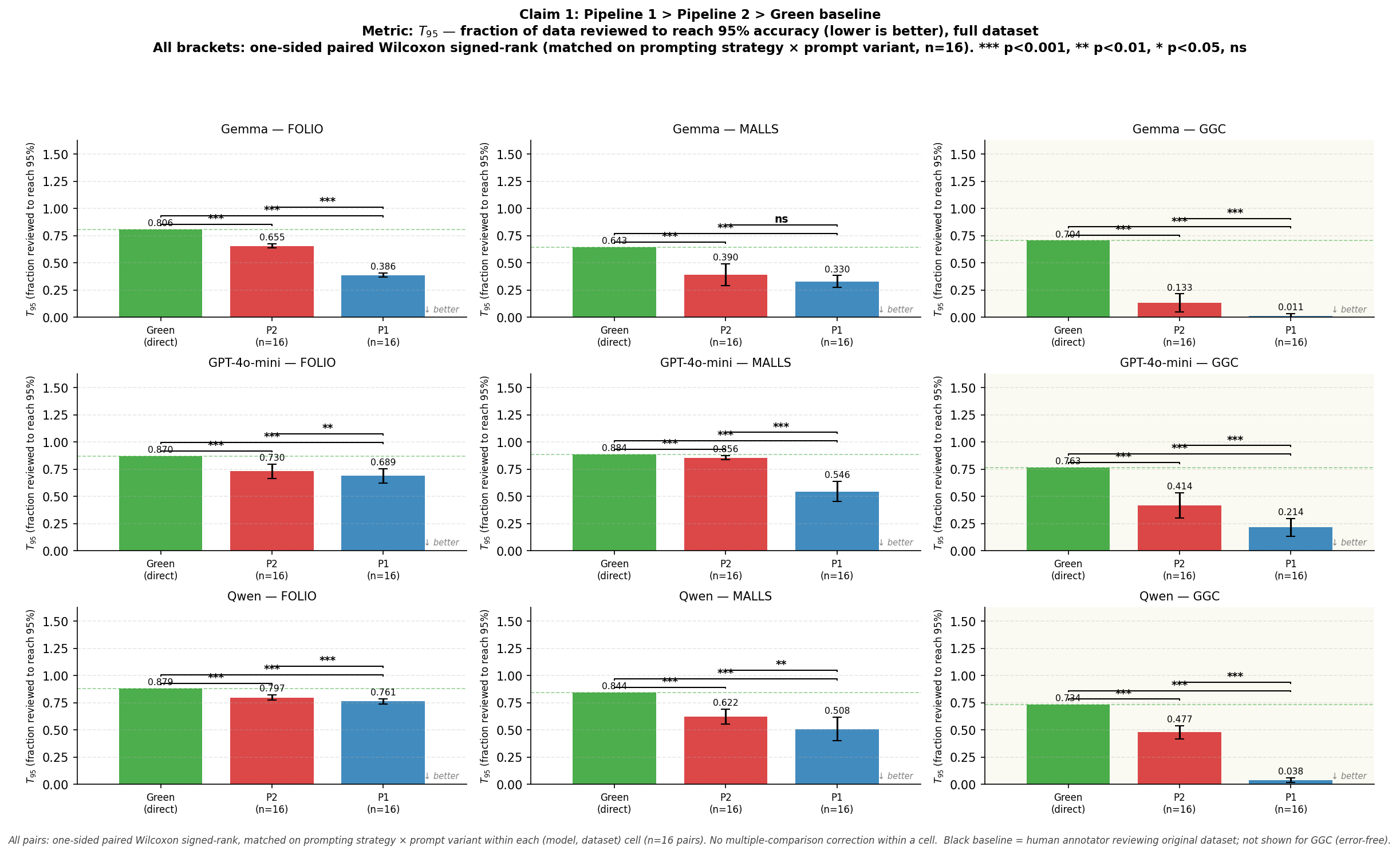}
    \caption{Pipeline comparison across models and datasets under the $T_{95}$ metric.}
    \label{fig:t95_pipeline_comparison}
\end{figure*}

\begin{figure*}
    \centering
    \includegraphics[width = \linewidth, trim={0 1cm 0 3cm},clip]{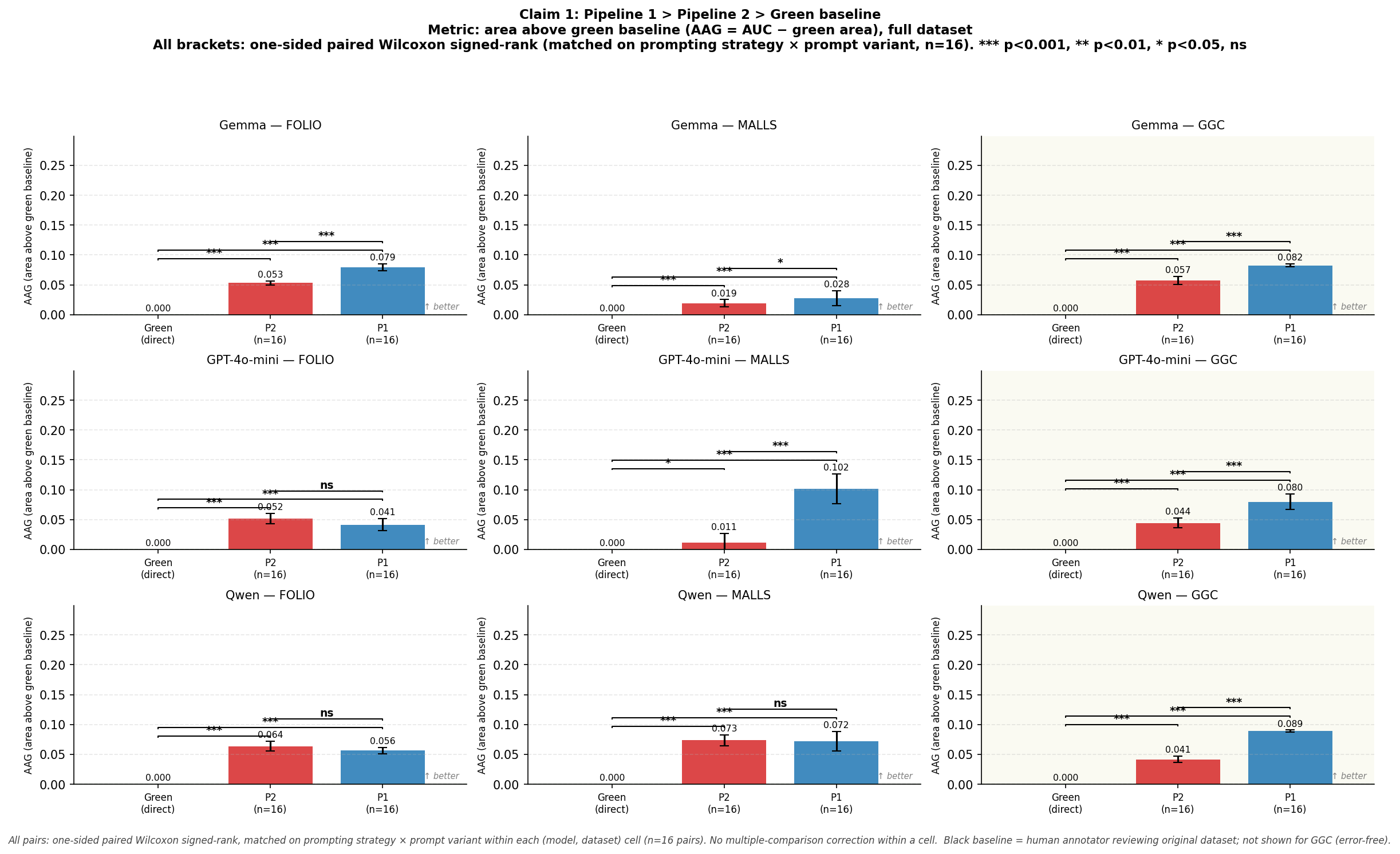}
    \caption{Pipeline comparison across models and datasets under the AAG metric.}
    \label{fig:aag_pipeline_comparison}
\end{figure*}

The main paper reports AUC (area under the curve) as the primary scalar summary of pipeline quality. Figure \ref{fig:t90_pipeline_comparison}-\ref{fig:t95_pipeline_comparison} replicate the same 3×3 bar grid (model × dataset) for the other complementary metrics $T_{90}$ (fraction of the dataset that must be reviewed to reach 90\% accuracy), and $T_{95}$ (fraction to reach 95\% accuracy). 

Figure \ref{fig:aag_pipeline_comparison} complements the analysis with the additional \emph{Area Above Green} (AAG) metric: the area of the region in the
accuracy--human effort plane that lies between the Green Baseline curve
and the curve of the strategy under evaluation. By construction, the
Green Baseline has AAG $= 0$, so every positive value directly
quantifies the net gain over starting the curation from the datasets obtained by the direct LLM translation.

All statistical tests use one-sided paired Wilcoxon signed-rank matched on prompting strategies × prompt variants within each cell ($n=16$ pairs); no multiple-comparison correction is applied within a cell.

 The $T_{90}$ and $T_{95}$ bars (Fig. \ref{fig:t90_pipeline_comparison}–\ref{fig:t95_pipeline_comparison}) confirm the AUC picture from a different angle: Pipeline~1 consistently requires reviewing a smaller fraction of the dataset to reach the accuracy thresholds, and the gains are statistically significant in almost all cells.

\section{Model Comparison on Pipeline 1}
\label{app:model_comparison}
\begin{figure*}
    \centering
    \includegraphics[width = \linewidth, trim={0 0 0 4cm},clip]{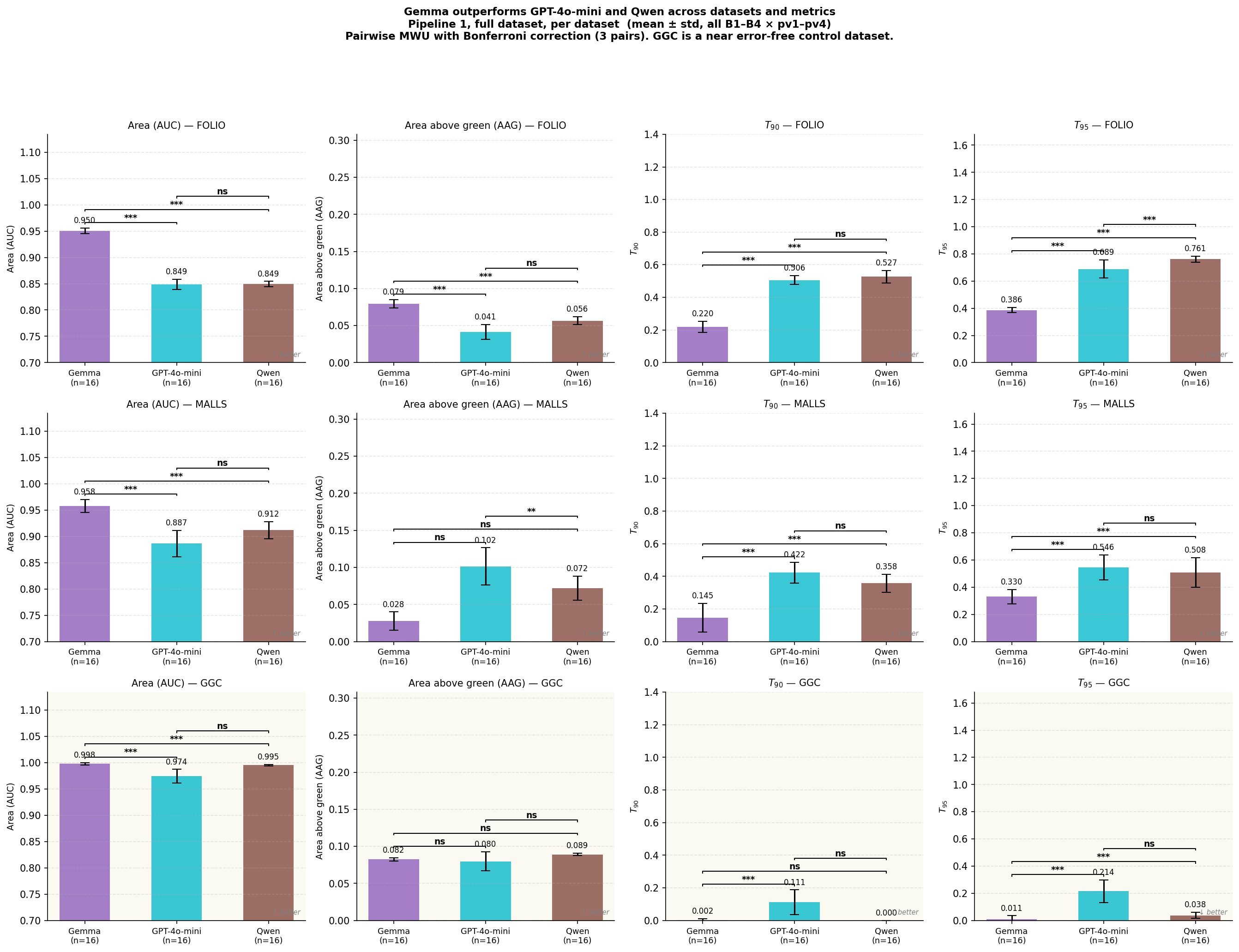}
    \caption{Model comparison across all three datasets and all four
    metrics (AUC, AAG, $T_{90}$, $T_{95}$) for Pipeline~1. Each bar
    represents the mean $\pm$ std over all 16 prompting combinations
    (B1--B4~$\times$~pv1--pv4). Brackets indicate pairwise
    Mann-Whitney U tests with Bonferroni correction (three pairs).}
    \label{fig:model_comparison}
\end{figure*}

Figure~\ref{fig:model_comparison} supports the model comparison
discussed in the main paper, showing results separately for all three
datasets and including the AAG metric (defined in
Appendix~\ref{app:pipeline_comparison}). Each bar represents the mean
$\pm$ std over all 16 prompting combinations (B1--B4~$\times$~pv1--pv4)
of Pipeline~1.

Gemma's advantage over the other models is consistent across
\textsc{FOLIO} and \textsc{MALLS}; on \textsc{GGC} the gap narrows
because all models operate near ceiling, as expected for an error-free
control dataset.

The AAG results require careful interpretation. On \textsc{MALLS} and
\textsc{GGC}, Gemma's AAG is lower than that of the other models,
which may seem inconsistent with its overall superiority. The key is that the Green Baseline is model-dependent: it is anchored to each model's own direct-translation accuracy. AAG therefore measures how much a pipeline improves over that model's own baseline, not over a common reference. A stronger model has a higher Green Baseline to
beat, so the same absolute accuracy gains translate into a smaller AAG.
This is a property of the metric, not a failure of the pipeline.

\section{Variant Selection and Sensitivity}
\label{app:variant_selection}
The main paper restricts the Pipeline~1 analysis to the subset
\{B1, B2, B3\}~$\times$~\{pv1, pv3\}, excluding B4 and pv2/pv4.
This appendix documents the statistical justification for that
restriction and shows that no analogous restriction is warranted for
Pipeline~2.

\begin{figure*}
    \centering
    \includegraphics[width = \linewidth, trim={0 1cm 0 2cm},clip]{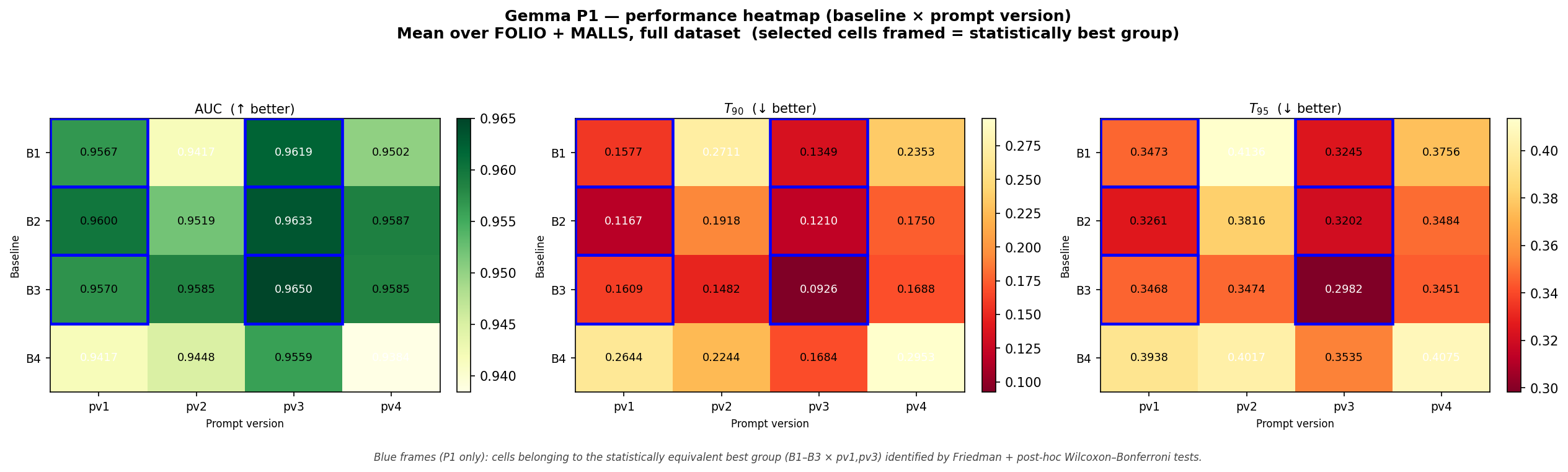}
    \caption{Prompting strategies and variants comparison for Pipeline 1. Blue-framed cells mark the retained group
    \{B1, B2, B3\}~$\times$~\{pv1, pv3\}.}
    \label{fig:heatmap_1}
\end{figure*}

\begin{figure*}
    \centering
    \includegraphics[width = \linewidth, trim={0 1cm 0 2cm},clip]{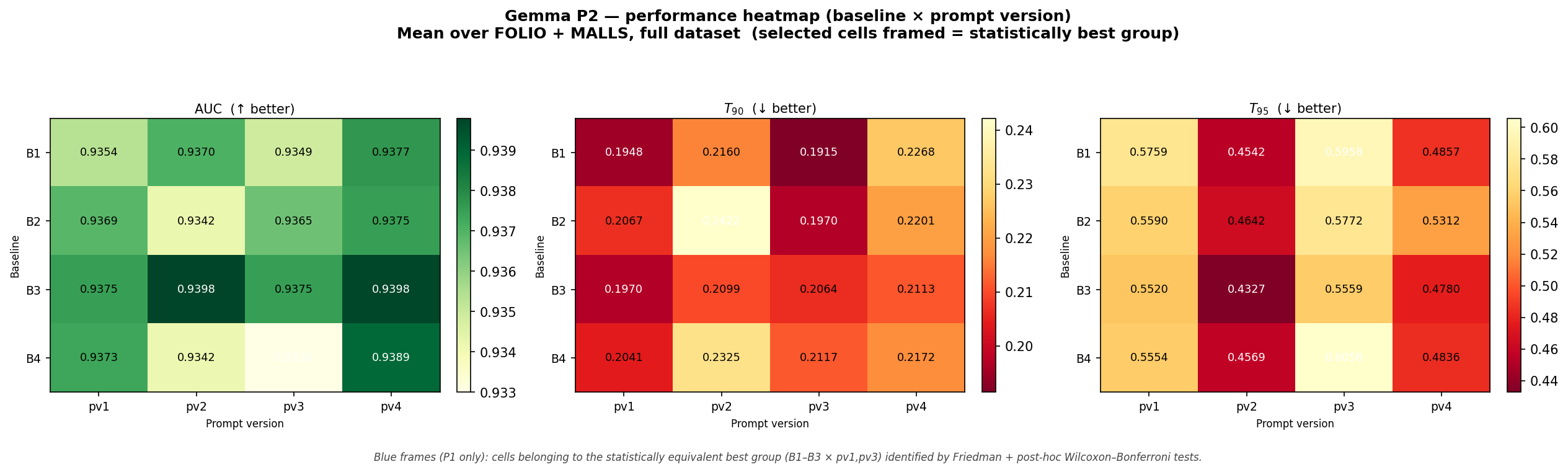}
    \caption{Prompting strategies and variants comparison for Pipeline 2}
    \label{fig:heatmap_2}
\end{figure*}

Figures \ref{fig:heatmap_1} and \ref{fig:heatmap_2} show  4$\times$4
heatmaps (prompting strategies $\times$ prompt variants) of AUC, $T_{90}$, and
$T_{95}$ for Gemma on Pipeline~1 and Pipeline~2, respectively,
averaged over \textsc{FOLIO} and \textsc{MALLS}.
In
Figure~\ref{fig:heatmap_1}, the blue-framed cells mark the retained
group, which is statistically equivalent within itself and superior to
the excluded configurations (B4 and pv2/pv4); higher (lighter) cell
values  indicate worse performance, confirming
that the statistical finding matches the visual pattern.

We use a two-stage non-parametric approach. In stage 1, a Friedman test (non-parametric repeated-measures ANOVA) tests the global null that all levels of a factor are exchangeable. The blocking structure is: for the prompt strategy factor, each block is one (dataset × prompt variant) combination (8 blocks, 4 levels); for the prompt-version factor, each block is one (dataset × prompting strategy) combination (8 blocks, 4 levels). In stage 2, all $\binom{4}{2}=6$ pairs are tested with two-sided paired Wilcoxon signed-rank; p-values are Bonferroni-corrected by multiplying by 6.

\paragraph{Pipeline 1.} The Friedman test is significant for both the prompting strategy factor (AUC: $\chi^2=14.55$, $p=0.002$; $T_{90}$: $p=0.001$; $T_{95}$: $p=0.002$) and the prompt variant factor (AUC: $\chi^2=11.25$, $p=0.010$; $T_{90}$: $p<0.001$; $T_{95}$: $p=0.003$). Post-hoc tests identify B4 as significantly inferior to B2 and B3 ($p_\text{bonf} \le 0.047$), and pv2 and pv4 as significantly inferior to pv1 ($p_\text{bonf} \le 0.047$). The pairs B1–B2, B1–B3, B2–B3 and the pairs pv1–pv3 are all non-significant, justifying retaining all three and both prompt variants in the best-combination group.

\paragraph{Pipeline 2.} The Friedman test is non-significant for both the prompting strategy factor ($\chi^2=1.85$, $p=0.583$) and the prompt variant factor ($\chi^2=0.15$, $p=0.985$), with all post-hoc pairs also non-significant. This indicates that the performance landscape for Pipeline 2 is essentially flat across prompting combinations: no restriction is statistically warranted. For the comparison in the main paper (Figure~\ref{fig:claim3}), we
therefore do not apply the \{B1, B2, B3\}~$\times$~\{pv1, pv3\}
restriction to Pipeline~2; instead, we select the configuration with
the highest AUC, which provides the strongest possible case for
Pipeline~2 and makes the superiority of Pipeline~1 all the more
notable.

\section{Oversight Band Grid and Best Variant Results}
\label{app:full-results}
\begin{figure*}
    \centering
    \includegraphics[width = \linewidth, trim={0 2cm 0 2cm},clip]{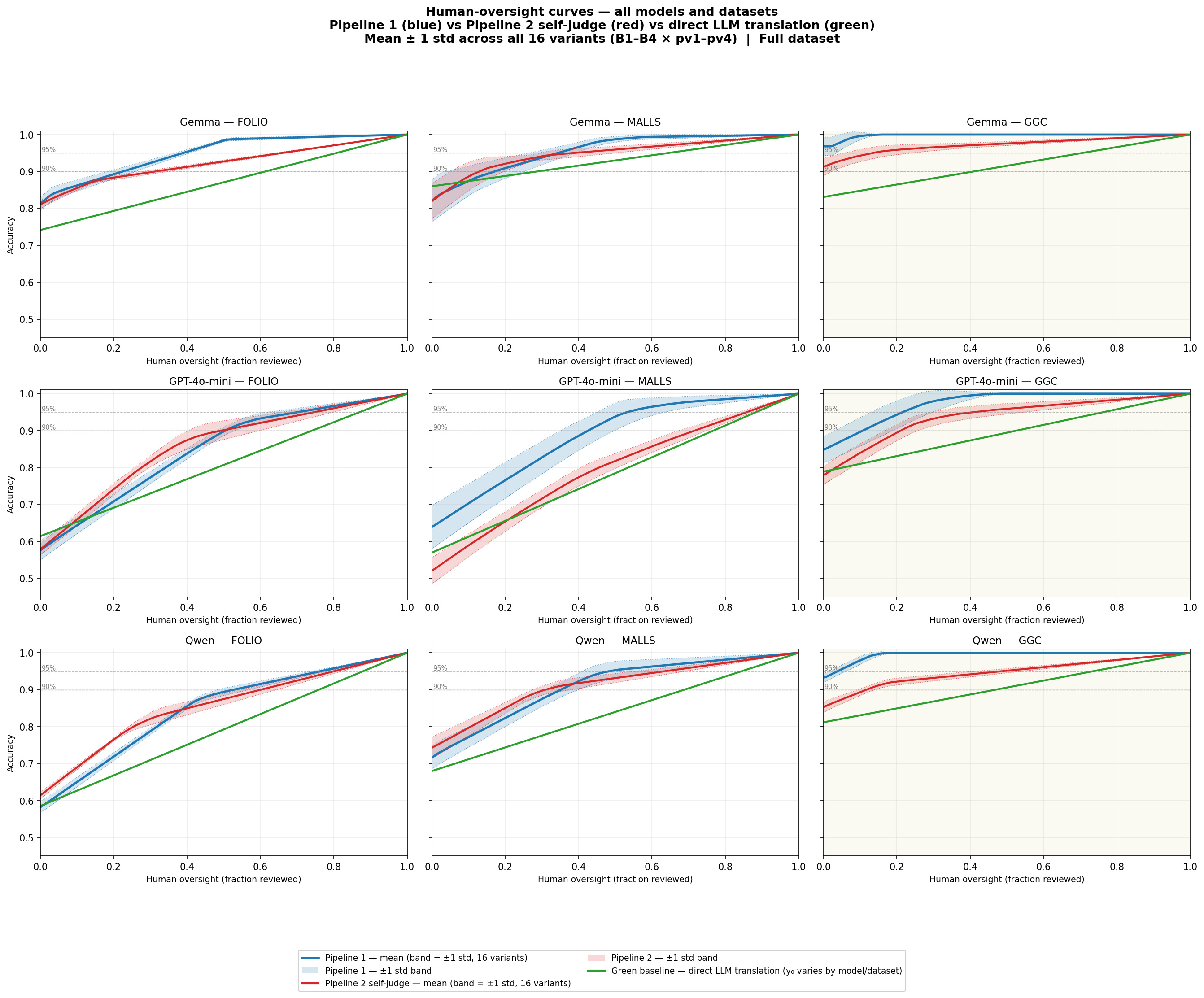}
    \caption{Extension of Figure \ref{fig:claim3} to all the models and datasets. Each band shows mean $\pm$ std over the 16 variants for Pipeline~1 (blue),
    and for Pipeline~2 (red); the green line represents the Green Baseline.}
    \label{fig:bandgrid}
\end{figure*}

Figure \ref{fig:bandgrid} shows the full 3$\times$3 grid of oversight curves (model$\times$dataset) for Pipeline 1 (blue band), Pipeline 2 (red band), and the direct-translation Green Baseline (green line). The figure generalizes the Gemma-only plot from the main paper (Fig. \ref{fig:claim3}) and supports three observations: \begin{itemize}
    \item \textbf{Ordering is universal.} The Pipeline~1 $>$ Pipeline~2
    $>$ Green Baseline ordering (by AUC) visible for Gemma holds for GPT-4o-mini and Qwen as well on FOLIO and MALLS. The relative gap between Pipeline~1 and Pipeline~2 is largest for Gemma.
    \item \textbf{\textsc{GGC} is near-ceiling for all models.} All three models with Pipeline 1 converge to $\geq$ 0.95 accuracy almost immediately. The oversight gain is real but compressed, confirming \textsc{GGC}'s role as an error-free control.
\end{itemize}

For completeness, Table~\ref{tab:best2_variants} reports, for each
model, dataset, and pipeline, the two prompt configurations with the
highest AUC, together with their scores under all metrics considered
(AUC, $T_{90}$, $T_{95}$). For comparison, we report also for each model and dataset the scores for the Black Baseline and the Green Baseline.

\begin{table*}[t]
\centering
\setlength{\tabcolsep}{5pt}
\small
\begin{tabular}{llcccccc}
\toprule
\textbf{Model} & \textbf{Dataset} & \textbf{Pipe} & \textbf{Strategy} & \textbf{Variant}
  & \textbf{AUC ($\uparrow$)}  & $\boldsymbol{T_{90}}$ ($\downarrow$) & $\boldsymbol{T_{95}}$ ($\downarrow$) \\
\midrule

\multirow{12}{*}{\textbf{Gemma}}
  & \multirow{6}{*}{FOLIO}
    & Black & - & - & 0.787 & 0.765 & 0.882 \\ &
    & Green & - & - & 0.871 & 0.613 & 0.806 \\ &
    & P\textsubscript{1} & B2 & pv3 & 0.959 & 0.169 & 0.352 \\
  &    & P\textsubscript{1} & B3 & pv2 & 0.958 & 0.176 & 0.360 \\
  &    & P\textsubscript{2} & B4 & pv2 & 0.931 & 0.228 & 0.614 \\
  &    & P\textsubscript{2} & B1 & pv2 & 0.929 & 0.234 & 0.617 \\
\cmidrule(l){2-8}
  & \multirow{6}{*}{MALLS}
    & Black & - & - & 0.790 & 0.762 & 0.881 \\ &
    & Green & - & - & 0.930 & 0.286 & 0.643 \\ &
    & P\textsubscript{1} & B3 & pv3 & 0.974 & 0.010 & 0.230 \\
  &    & P\textsubscript{1} & B1 & pv3 & 0.969 & 0.080 & 0.280 \\
  &    & P\textsubscript{2} & B1 & pv4 & 0.957 & 0.094 & 0.292 \\
  &    & P\textsubscript{2} & B3 & pv4 & 0.957 & 0.094 & 0.292 \\
\cmidrule(l){2-8}
  & \multirow{6}{*}{GGC}
    & Black & - & - & 1.000 & 0.000 & 0.000 \\ &
    & Green & - & - & 0.915 & 0.408 & 0.704 \\ &
    & P\textsubscript{1} & B2 & pv1 & 0.999 & 0.000 & 0.000 \\
  &    & P\textsubscript{1} & B3 & pv1 & 0.999 & 0.000 & 0.000 \\
  &    & P\textsubscript{2} & B2 & pv2 & 0.980 & 0.000 & 0.047 \\
  &    & P\textsubscript{2} & B3 & pv4 & 0.979 & 0.000 & 0.064 \\
\midrule

\multirow{12}{*}{\textbf{GPT-4o-mini}}
  & \multirow{6}{*}{FOLIO}
    & Black & - & - & 0.787 & 0.765 & 0.882 \\ &
    & Green & - & - & 0.807 & 0.741 & 0.870 \\ &
    & P\textsubscript{1} & B3 & pv1 & 0.873 & 0.464 & 0.548 \\
  &    & P\textsubscript{1} & B2 & pv1 & 0.861 & 0.499 & 0.584 \\
  &    & P\textsubscript{2} & B3 & pv1 & 0.873 & 0.400 & 0.700 \\
  &    & P\textsubscript{2} & B3 & pv2 & 0.869 & 0.453 & 0.727 \\
\cmidrule(l){2-8}
  & \multirow{6}{*}{MALLS}
    & Black & - & - & 0.790 & 0.762 & 0.881 \\ &
    & Green & - & - & 0.785 & 0.767 & 0.884 \\ &
    & P\textsubscript{1} & B3 & pv2 & 0.910 & 0.367 & 0.447 \\
  &    & P\textsubscript{1} & B1 & pv1 & 0.910 & 0.376 & 0.462 \\
  &    & P\textsubscript{2} & B3 & pv2 & 0.830 & 0.653 & 0.827 \\
  &    & P\textsubscript{2} & B2 & pv2 & 0.825 & 0.662 & 0.831 \\
\cmidrule(l){2-8}
  & \multirow{6}{*}{GGC}
    & Black & - & - & 1.000 & 0.000 & 0.000 \\ &
    & Green & - & - & 0.895 & 0.526 & 0.763 \\ &
    & P\textsubscript{1} & B4 & pv1 & 0.993 & 0.000 & 0.072 \\
  &    & P\textsubscript{1} & B4 & pv2 & 0.990 & 0.000 & 0.096 \\
  &    & P\textsubscript{2} & B2 & pv3 & 0.950 & 0.170 & 0.281 \\
  &    & P\textsubscript{2} & B1 & pv3 & 0.949 & 0.173 & 0.255 \\
\midrule

\multirow{12}{*}{\textbf{Qwen}}
  & \multirow{6}{*}{FOLIO}
    & Black & - & - & 0.787 & 0.765 & 0.882 \\ &
    & Green & - & - & 0.793 & 0.759 & 0.879 \\ &
    & P\textsubscript{1} & B1 & pv3 & 0.862 & 0.460 & 0.719 \\
  &    & P\textsubscript{1} & B1 & pv1 & 0.855 & 0.506 & 0.753 \\
  &    & P\textsubscript{2} & B4 & pv4 & 0.868 & 0.510 & 0.755 \\
  &    & P\textsubscript{2} & B1 & pv4 & 0.867 & 0.486 & 0.743 \\
\cmidrule(l){2-8}
  & \multirow{6}{*}{MALLS}
    & Black & - & - & 0.790 & 0.762 & 0.881 \\ &
    & Green & - & - & 0.840 & 0.688 & 0.844 \\ &
    & P\textsubscript{1} & B4 & pv4 & 0.935 & 0.304 & 0.413 \\
  &    & P\textsubscript{1} & B1 & pv3 & 0.934 & 0.301 & 0.395 \\
  &    & P\textsubscript{2} & B1 & pv3 & 0.922 & 0.265 & 0.562 \\
  &    & P\textsubscript{2} & B3 & pv3 & 0.921 & 0.285 & 0.529 \\
\cmidrule(l){2-8}
  & \multirow{6}{*}{GGC}
    & Black & - & - & 1.000 & 0.000 & 0.000 \\ &
    & Green & - & - & 0.906 & 0.468 & 0.734 \\ &
    & P\textsubscript{1} & B4 & pv3 & 0.997 & 0.000 & 0.016 \\
  &    & P\textsubscript{1} & B2 & pv3 & 0.997 & 0.000 & 0.003 \\
  &    & P\textsubscript{2} & B3 & pv4 & 0.955 & 0.083 & 0.393 \\
  &    & P\textsubscript{2} & B2 & pv3 & 0.954 & 0.097 & 0.393 \\

\bottomrule
\end{tabular}
\caption{Table representing the AUC, $T_{90}$, and $T_{95}$ of the best 2 variants (according to AUC) for each model, dataset and pipeline}
\label{tab:best2_variants}
\end{table*}


\end{document}